\documentclass[10pt,journal,compsoc]{IEEEtran}
\ifCLASSOPTIONcompsoc
  \usepackage[nocompress]{cite}
\else
  \usepackage{cite}
\fi
\ifCLASSINFOpdf
\else
\fi
\usepackage{amsmath}
\usepackage{array}

\usepackage{times}
\usepackage{epsfig}
\usepackage{graphicx}
\usepackage{amssymb}

\usepackage{bm} 
\usepackage{multirow}
\usepackage{float}

\usepackage{subfig}
\usepackage{epsfig}
\usepackage{xcolor}
\usepackage[linesnumbered,ruled]{algorithm2e}
\usepackage{siunitx}
\usepackage{booktabs}
\usepackage{mathtools}
\usepackage{vcell}
\usepackage{enumitem}

\usepackage{hyperref}
\hypersetup{
    colorlinks=true,
    linkcolor=blue,
    filecolor=magenta,      
    urlcolor=cyan,
}

\newcommand\todo[1]{\textcolor{red}{#1}}

\begin{document}
%
\title{Time-Ordered Recent Event (TORE) Volumes for Event Cameras}

\author{R.~Wes~Baldwin,~\IEEEmembership{Member,~IEEE,}
        ~Ruixu~Liu,~\IEEEmembership{Member,~IEEE,}
        ~Mohammed~Almatrafi,~\IEEEmembership{Member,~IEEE,}
        ~Vijayan~Asari,~\IEEEmembership{Senior~Member,~IEEE}
        and ~Keigo~Hirakawa,~\IEEEmembership{Senior~Member,~IEEE}
\IEEEcompsocitemizethanks{\IEEEcompsocthanksitem R. Baldwin, R. Liu, V. Asari, and K. Hirakawa are with the Department of Electrical and Computer Engineering, University of Dayton, Dayton, US, 45469. \protect \\
Email: (baldwinr2,liur05,vasari1,khirakawa1)@udayton.edu.
\IEEEcompsocthanksitem M. Almatrafi is with the Department
of Electronic and Communication Engineering, Umm Al-Qura University, Al-Lith,
Saudi Arabia, 28434. \protect\\
E-mail: mmmatrafi@uqu.edu.sa.}
}

%
%

\markboth{Journal of \LaTeX\ Class Files,~Vol.~14, No.~8, August~2015}%
{Shell \MakeLowercase{\textit{et al.}}: Bare Demo of IEEEtran.cls for Computer Society Journals}
%



\IEEEtitleabstractindextext{%
\begin{abstract}
Event cameras are an exciting, new sensor modality enabling high-speed imaging with extremely low-latency and wide dynamic range. Unfortunately, most machine learning architectures are not designed to directly handle sparse data, like that generated from event cameras. Many state-of-the-art algorithms for event cameras rely on interpolated event representations---obscuring crucial timing information, increasing the data volume, and limiting overall network performance. This paper details an event representation called Time-Ordered Recent Event (TORE) volumes. TORE volumes are designed to compactly store raw spike timing information with minimal information loss. This bio-inspired design is memory efficient, computationally fast, avoids time-blocking (i.e. fixed and predefined frame rates), and contains ``local memory'' from past data. The design is evaluated on a wide range of challenging tasks (e.g. event denoising, image reconstruction, classification, and human pose estimation) and is shown to dramatically improve state-of-the-art performance. TORE volumes are an easy-to-implement replacement for any algorithm currently utilizing event representations.

\end{abstract}

}

\maketitle

\IEEEdisplaynontitleabstractindextext

%
\IEEEpeerreviewmaketitle

\IEEEraisesectionheading{\section{Introduction}\label{sec:introduction}}

%
%
%
%

\IEEEPARstart{E}{vent-based} neuromorphic cameras (a.k.a. ``event cameras'') are an emerging technology that has the potential to significantly alter the future of many computer vision applications. The Dynamic Vision Sensor (DVS) in event cameras is a so-called \emph{silicon retina}, loosely emulating the changes that take place in the membrane potential of biological photoreceptors. Specifically, the DVS encodes each pixel's log-intensity changes as ``events,'' mimicking the spiking neurotransmission within biological visual pathways. The events at each pixel are reported asynchronously and are time-stamped with microsecond-order latency and resolution. Event cameras also have a far wider dynamic range than conventional cameras, allowing them to see extremely bright and shadowed areas simultaneously. Additionally, event cameras generate significantly less data volume than standard high-speed systems due to the sparse nature of the data. All of these factors make event cameras exceptional for real-time and edge-based computing with significant benefits for speed and power consumption.

Sparse event data are well-matched for spiking neural networks (SNN), which process asynchronous data efficiently. Event cameras have grown rapidly in resolution over the last several years, but commercially available SNNs have not yet matched this growth. Modern deep learning architectures, such as the state-of-the-art convolutional neural networks (CNN) implemented on modern GPUs and CPUs, can easily handle event camera data volume rates. However, this type of data can be challenging to process in a synchronous manner. This challenge is two-fold. First, asynchronous event data must be transformed into an intermediary representation that is ingestible by synchronous deep learning architectures. This mapping of asynchronous (i.e. sparse) camera events to synchronous (i.e. dense) framing data is irreversible and, therefore, lossy. Thus, it is desirable to construct a synchronous representation of asynchronous events that best preserve the spatial-temporal integrity of the event data. Second, machine learning architectures, based on spatial convolutions, are not yet fully optimized for sparse data input; however, it is an active area of research~\cite{hamilton2017inductive,qi2017pointnet,qi2017pointnet++,velivckovic2017graph,zhou2018voxelnet}. Convolution and other linear operators are effective at sparsifying dense data by exploiting redundancies among neighboring pixels, but they often have the opposite effect when the data is already sparse. 

\begin{figure}[t]
\centering
\begin{tabular}{m{0.45\linewidth}m{0.45\linewidth}}
\includegraphics[width=1.06\linewidth]{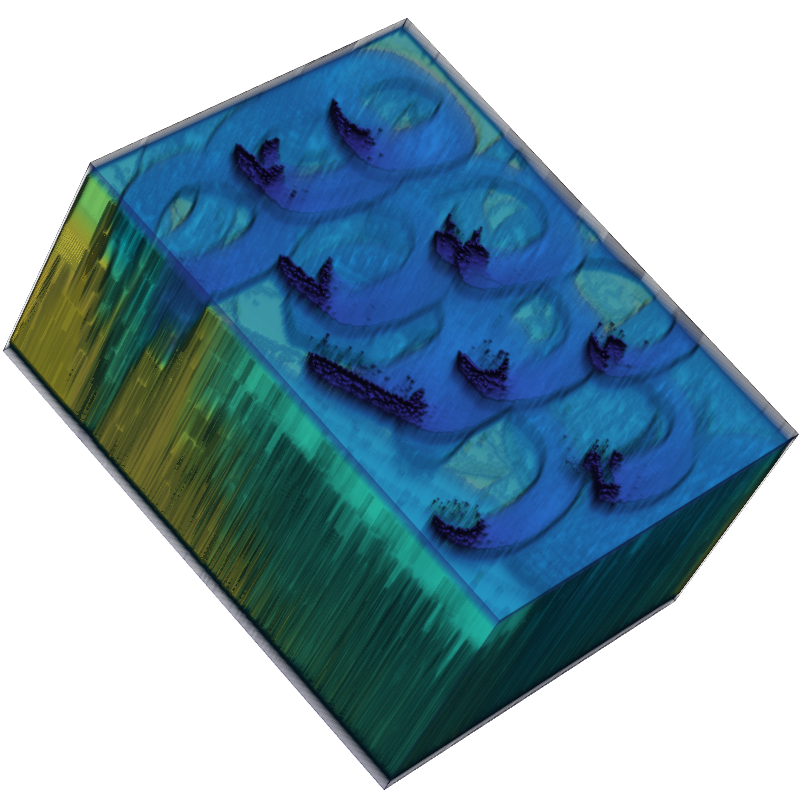} & \includegraphics[width=1.06\linewidth]{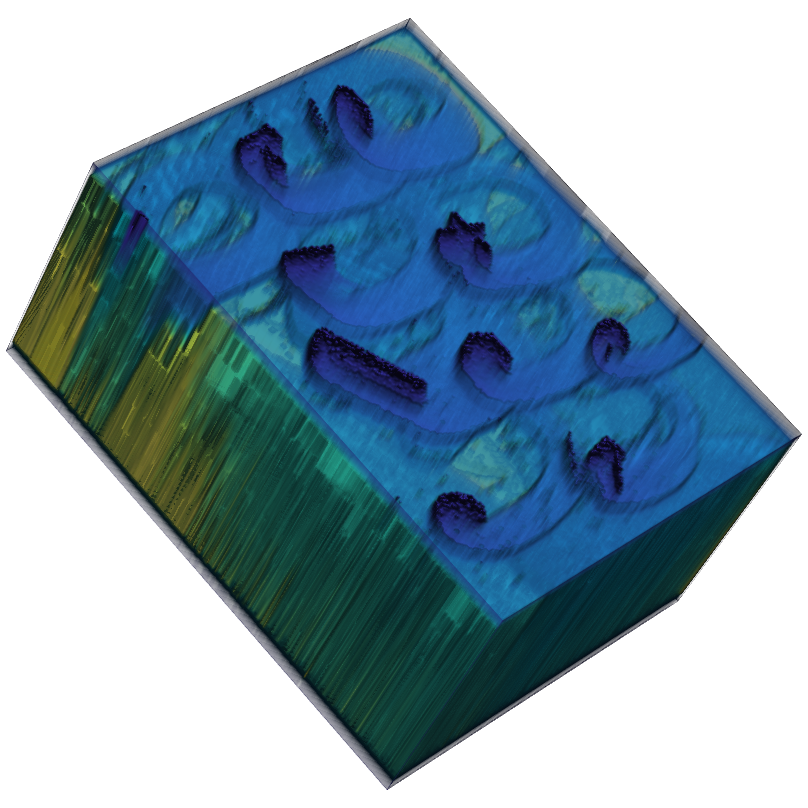}\\
\multicolumn{1}{c}{Positive Events} & \multicolumn{1}{c}{Negative Events}
\end{tabular}
\caption{TORE volume example generated from \emph{shapes\_6dof} in the Event Camera Dataset~\cite{mueggler2017event}. TORE volumes are a dense event representation that encodes a history of recent events.}
\label{fig:eyecandy}
\end{figure}

\emph{Synchronous event data representation}, described above, is an often overlooked area of research. While the majority of the event camera work is focused on the novelties of the application and the machine learning architectures designed for event detection, we emphasize in our paper that the design of synchronous event representations influence algorithm performance in the application domain. To this effect, we develop a compact and memory efficient transformation of event data to synchronous frames that captures the key spatial-temporal evolution of events that is most pertinent to computer vision applications. We summarize our contributions below:
\begin{itemize}[leftmargin=5.5mm]
  \item {\bf Time-Ordered Recent Events (TORE) Volume:} Minimally-lossy, bio-inspired synchronous event representation for event camera data that support low- and high-level tasks;
  \item {\bf Event Denoising:} State-of-the-art performance for event denoising on real-world data via a localized implementation.
  \item {\bf Image Reconstruction:} A straightforward, adaptable, and effective method to convert event camera output to high-speed video trained on real-world data.
  \item {\bf Event-based Classification:} State-of-the-art performance for event-based object classification across multiple datsets.
  \item {\bf 2D and 3D Pose Estimation:} State-of-the-art performance for 2D and 3D human pose estimation via event cameras.
\end{itemize}

The rest of this paper is organized as follows. The remainder of Section~\ref{sec:introduction} presents background on event cameras. In Section~\ref{sec:eventRepresentations}, prior work in event representations is reviewed and inspiration for a new representation is drawn from the human retina. Section~\ref{sec:tore} details the proposed representation design and outlines multiple benefits over existing alternatives. Sections~\ref{sec:denoise}--\ref{sec:pose} test various applications of the proposed design against diverse datasets and report results. Conclusions are drawn in Section~\ref{sec:conclusion}.

Preliminary results of this work appeared in our paper on event denoising~\cite{baldwin2020event}. This paper builds upon that work by generalizing the TORE event representations and testing performance on a wide-range of tasks.

\subsection{Background: Event Cameras}

\begin{figure}[tbp]
    \centerline{\includegraphics[width=1.0\linewidth]{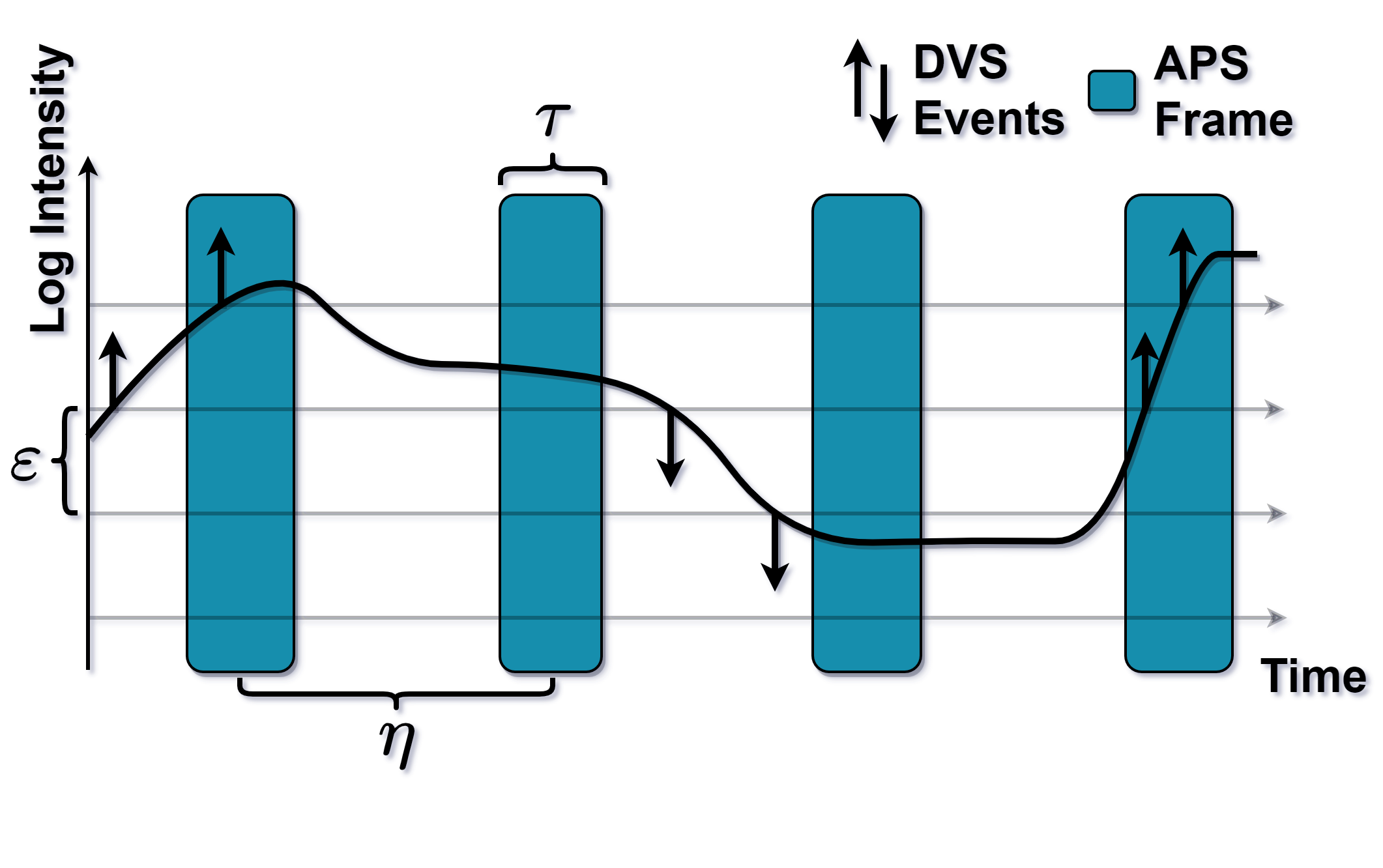}}
    \caption{The continuous log-intensity $J$ for a single pixel within an event camera is shown as the solid black line. DVS events are generated for this pixel when the log-intensity exceeds a predefined threshold $\varepsilon$. In certain event cameras, APS frames are also generated using the same photodiode. APS frames are exposed for $\tau$ seconds, occurring once every $\eta$ seconds.}
    \label{fig:eventGen}
\end{figure}

Neuromorphic cameras generate events whenever a log-intensity change is detected within a pixel. For example, let $J(x,y,t)$ denote the log-intensity at pixel $(x,y)\in\mathbb{Z}^2$ and at time $t\in\mathbb{R}$. Each event $\bm{e}=(e_x,e_y,e_t,e_p)$ reports the spatial location $(x,y)=(e_x,e_y)$, timestamp $t=e_t$ that $J(x,y,t)$ exceeds the threshold, and polarity $e_p\in\{+1,-1\}$, indicating whether the change in the log-intensity was a result from an increase or decrease in intensity respectively (see Figure~\ref{fig:eventGen}). In an ideal case, events are triggered when the log-intensity exceeds a predefined threshold set within the camera hardware $\varepsilon$. Mathematically, the $i$th event $\bm{e}^i=(e^i_x,e^i_y,e^i_t,e^i_p)$ is defined as
\begin{align}
\begin{split}
    (e^i_x,e^i_y,e^i_t) &= \arg \min_{(x,y,t)} \\
    &\left\{t>e^{i-1}_t\Big| |J(x,y,t)-R(x,y)|\geq \varepsilon\right\}\\
    e^i_p &= \operatorname{sign}\left(J(e^i_x,e^i_y,e^i_t)-R(e^i_x,e^i_y)\right),
\end{split}
\end{align}
where $\bm{e}^{i-1}=(e^{i-1}_x,e^{i-1}_y,e^{i-1}_t,e^{i-1}_p)$ is the sensor's previous event, and $R(x,y)$ is the reference value at pixel $(x,y)$ based on the log-intensity value of the most recent event.

In addition to events from the DVS, some neuromorphic cameras provide a standard framing video readout from the same photodiode. This video stream is called an Active Pixel Sensor (APS) and is generated with the typical integrate and readout pattern of a standard camera. APS frames are generated at $\sim30$Hz and have the same per-pixel field-of-view as the DVS events as they both derive their signal from a shared photodiode. APS frame data is also time-synchronized with the DVS events, making it extremely useful for DVS algorithm evaluation and scene situational awareness. APS frames are shown in Figure~\ref{fig:eventGen} as vertical bars with the width and spacing of the bar representing the integration interval ($\tau$) and the frame periodicity ($\eta$), respectively.

\section{Event Representations}\label{sec:eventRepresentations}

Owing to the asynchronous nature of data, processing steps in event-based computer vision are considerably different from conventional framing cameras. Event representation is the first step in the event-based computer vision architecture, where the raw DVS data is transformed into a spatial-temporal format interpretable by the subsequent processing steps (such as deep neural networks) that carry out tasks such as classification, motion tracking, prediction, image reconstruction, and 6 DoF estimation.


One may categorize event representation roughly into four modalities. First modality is the spike processing, such as SNN, that natively support sparse asynchronous data~\cite{akopyan2015truenorth}. While efficient in power consumption and effective in methodology, machine learning in spiking forms are less flexible and more difficult to train than traditional CNNs. They also require specialized hardware not available in most computational platforms. Second modality is an analytical event processing, where the timing and the polarity of the events are explicitly exploited. Examples include local plane fitting \cite{benosman2013event} and temporal derivative interpolation \cite{almatrafi2019davis}. Most analytical event representations are task-specific, however, and thus do not generalize to a wide range of applications easily. Third modality is an ``intermediary representation'' of event camera data---which is the subject of this paper---to be paired with machine learning methods in synchronous form. To leverage the rapid advancements in CNNs and their recent successes in computer vision, event data needs to be transformed into a proxy 2D image-like or 3D video frame-like representation we call ``proxy frames.'' While it is impossible to losslessly represent asynchronous event data using only a few proxy frames, attention is placed on the representation that best captures the spatial-temporal evolution of events at each pixel. Fourth modality is intensity image reconstruction, which may be thought of as the special case of intermediary representation. By estimating intensity values for each pixel using events, machine learning methods trained on conventional images can be directly utilized~\cite{rebecq2019events}. 


\begin{table*}[htbp]
    \centering
    \caption{Comparison of various event representation methods used in machine learning for event-based camera data. \emph{H} and \emph{W} denote representation height and width, respectively. This table is expanded from \cite{gehrig2019end} to include new methods.}
    \resizebox{\textwidth}{!}{
    \begin{tabular}{lcll} 
        \toprule
        Event Representation & Dimensions & Description & Characteristics \\
        \midrule
        Event Frame~\cite{rebecq2017real} & $\emph{H}\times\emph{W}$ & Image of event polarities & Discards temporal and polarity information \\
        
        Distance Transform~\cite{almatrafi2020distance} & $\emph{H}\times\emph{W}$ & Image of spatial distance to active pixel & Discards temporal and polarity information \\
        
        Graph-based~\cite{bi2019graph} & $\emph{U}\times\emph{V}$ & Graph of edges and vertices & Discards temporal and spatial information\\
        
        Event Count~\cite{maqueda2018event,zhu2018ev} & $2\times\emph{H}\times\emph{W}$ & Image of event counts & Discards temporal information \\
        
        Surface of Active Events (SAE)~\cite{benosman2013event} & $2\times\emph{H}\times\emph{W}$ & Image of most recent timestamp & Discards all prior time stamps \\
        
        Voxel Grid~\cite{zhu2019unsupervised} & $\emph{B}\times\emph{H}\times\emph{W}$ & Voxel grid summing event polarities & Discards polarity information \\
        
        Averaged Time Surfaces~\cite{sironi2018hats} & $2\times\emph{H}\times\emph{W}$ & Image of average timestamp for window & Discards temporal information \\
        
        Inceptive Time Surfaces~\cite{baldwin2019inceptive} & $3\times\emph{H}\times\emph{W}$ & Image of filtered timestamps {\&} event count & Discards temporal information \\
        
        Event Spike Tensor (EST)~\cite{gehrig2019end} & $2\times\emph{B}\times\emph{H}\times\emph{W}$ & 4D grid of convolutions & Temporally quantizes information into \emph{B} bins \\
        
        \textbf{TORE Volumes}~\cite{baldwin2020event} & $2\times\emph{K}\times\emph{H}\times\emph{W}$ & 4D grid of last K timestamps & Retains all information for last \emph{K} events \\
        
        \bottomrule
    \end{tabular}}
\label{tab:representations}
\end{table*}

Table~\ref{tab:representations} outlines several published techniques and documents the rapid growth in representation complexity, where $H$ and $W$ are input height and width, respectively. The majority of these methods are optimized for high-level applications such as object detection, image reconstruction, or optical flow. In these use cases, it is common to use large representations where $H$ and $W$ are the sensor height and widths. By contrast, smaller spatial patches of events are used in low-level or event-level applications such as denoising or feature tracking. In this case, $H$ and $W$ represent the height and width of the patches centered around the event of interest, significantly smaller than the sensor height/width. Smaller size is requisite since representations are now generated per event and should require minimal processing. 

Among the synchronous intermediary representation of events, the majority of methods incorporate temporal windowing to group incoming events. Windowing may be a fixed interval to yield a constant frame rate proxy frames or may be a fixed event number yielding adaptive frame rate (sometimes referred to as ``event windowing''). Events occurring within a given temporal window are summarized to yield a proxy frame using one of the following techniques. In one representation called ``event accumulation'' or ``event counting,'' events within a temporal window are separated by polarity and events are counted for each polarity type to yield two images of size $H\times W$~\cite{maqueda2018event,zhu2018ev}. While this approach retains polarity information, temporal information is lost. 

A similar approach is ``event frames'' or ``polarity summing,'' computed by summing the polarity of events within a given time window~\cite{rebecq2017real}. Appealing to the observation that higher number of positive or negative events corresponds to a greater intensity change (i.e.~edge), polarity summing yields a single $\emph{H}\times\emph{W}$ proxy frame approximating the overall intensity changes that occurred within the given temporal window. In another representation called distance surfaces~\cite{almatrafi2020distance}, pixel values of the proxy frame encode the distance to the event closest to that pixel. It was shown that distance surfaces satisfy the optical flow equation exactly (unlike optical flow equation in APS, which is only approximate), making this representation an attractive choice for estimating pixel motion.

There are also variations in the ways in which a temporal window is used. A representation called ``voxel grid'' takes a temporal quantization approach, using bilinear sampling kernel to map the events to the nearest temporal grid~\cite{zhu2019unsupervised}; although, information is lost whenever multiple events are mapped to the same quantized pixel. Event Spike Tensor (EST)~\cite{gehrig2019end} is a variation to the voxel grid in~\cite{zhu2019unsupervised}, separating in polarity and allow for any function to execute per voxel instead of just summing polarity information. A downside to a temporal quantization approach is that it is not invariant to timing, meaning a change in initialization time can yield different representations for the same dataset.

Time surface is an alternative to event accumulation and polarity summation representation that prioritizes event timing over edge magnitude. Time surface is a two dimensional representation that encodes time stamps as pixel values. Surface of Active Events (SAE)~\cite{benosman2013event} time surfaces retain the timing for the latest event at each pixel location (for each polarity). This is not ideal for slow motion sequences or imaging in low light environments, as event timing becomes ambiguous or can be interrupted by noise. One can increase robustness by leveraging the fact that a single edge generates multiple events. For instance, averaged time surface---where each pixel value reports the average event time of events generated within a temporal window~\cite{sironi2018hats}---reduces the impact of a noise occurring in isolation. However, averaging introduces ambiguity in reported event timing. Inceptive time surface~\cite{baldwin2019inceptive} is a hybrid approach drawing on time surface and polarity summation. Inceptive surface filtering is designed to remove events that do not occur in groups (as a single edge is expected to generate multiple events) and reports the first event timing (corresponding to the edge arrival timing) and the number of events occurring in such groups (corresponding to edge magnitude). This filters isolated events (i.e. noise), retains higher fidelity of temporal information as well as edge magnitude, but at the sacrifice of low contrast changes that will be filtered out. Inceptive surfaces are also non-causal, introducing latency in real-time applications.

Other approaches use compact graph-based representations~\cite{bi2019graph,bi2019graphst}. Graph-based methods transform events within a temporal window into a set of connected nodes, $\emph{U}\times\emph{V}$. Such compact representation allows for a reduction in compute and memory resources, while remaining true to the sparse nature of event cameras. However, constructing the graphs can be computationally intensive. Additionally, this representation discards fine temporal and spatial information during the transformation.

Some event representations are inspired by the biological functions of human vision and appeal to the observation that the visual cortex prioritizes the most recent stimuli and retain the last event timing in a form of a time surface~\cite{benosman2013event,lagorce2016hots,afshar2020event}. Similarly, modeled after the behavior of Leaky Integrate and Fire (LIF) neurons, events are treated as spikes that decay over time exponentially in magnitude~\cite{fourcaud2003spike}. Sums of these exponentially decaying spikes are sampled at fixed time intervals, giving rise to an event representation where more recent events have greater contributions.


A general issue with time-windowed or event-windowed representations is that they induce latency. For temporal windowed representations with fixed frame rate, events that occur early in the window are not evaluated until the time window completes. Event windowing is computationally more efficient, but they have no upper bound for delay since the event window is based solely on the event rate of the sensor. Windowed methods typically require predefined window sizes that cannot be changed without network retraining.


\subsection{Retina and Event Camera Comparison}\label{sec:intro_retina}

The design of the event camera was inspired by the retina, and they share some similarities. Both the human retina and event cameras rely on change in light, rather than absolute levels, to encode information from visual scenes. Event cameras are bioinspired in that they generate a sparse signal only when change is detected. This is similar to how photoreceptors in the human eye function, but prior to reaching the optic nerve, photoreceptor output is processed by the retina, which converts the electrical activity of each photoreceptor into an action potential (i.e. sparse cell-to-cell communication).

\begin{figure}[htbp]
    \centerline{\includegraphics[width=1.0\linewidth]{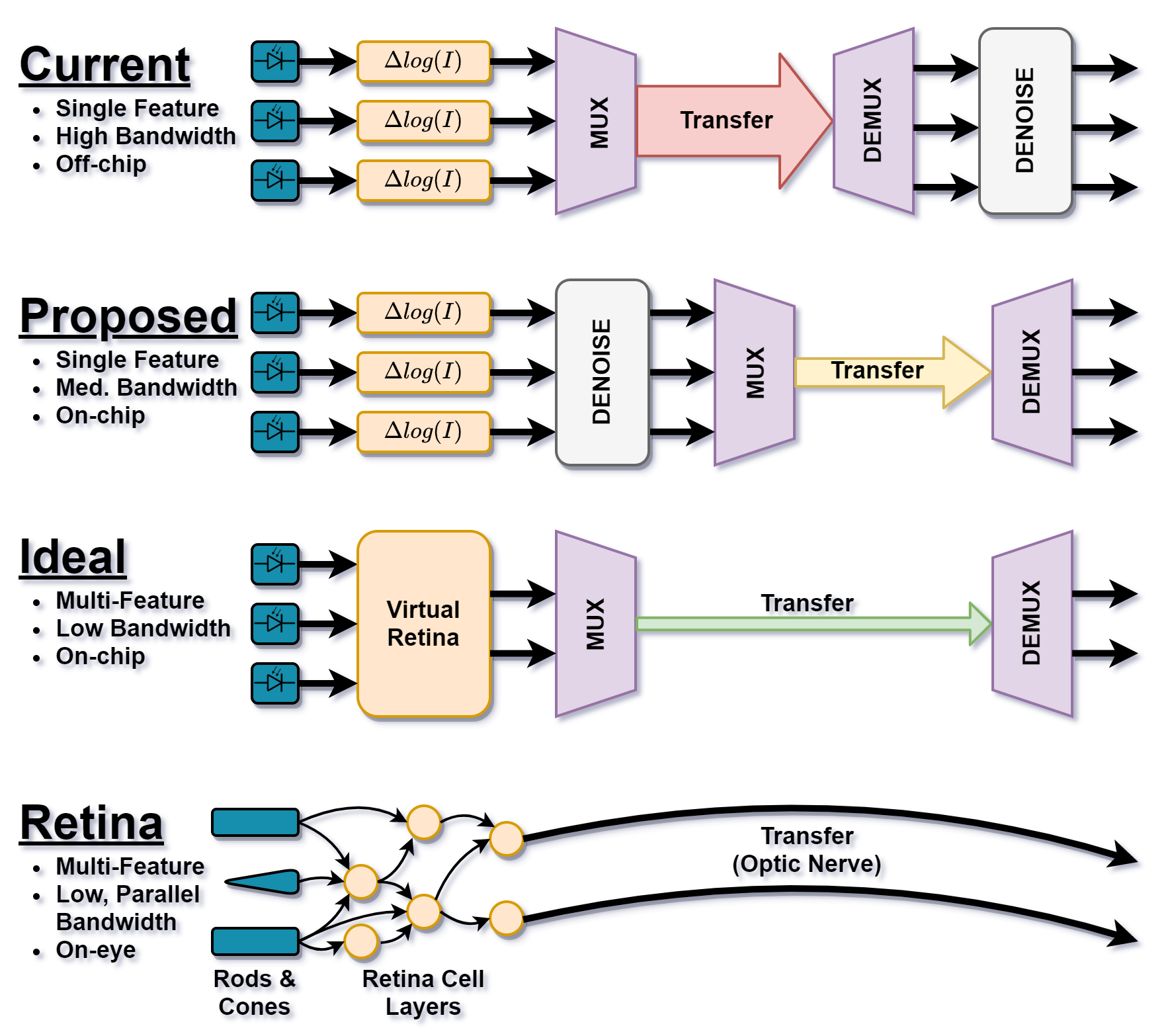}}
    \caption{Current denoising is performed after data has been read out and transmitted from the sensor. This makes the sensor vulnerable to noise saturating the data link and requires downstream application to process more data. This work develops a method using only local data, similar to the retina, to denoise event data. This method could even be moved ``on-chip'' and incorporated into a full virtual retina.}
    \label{fig:nowNextGoal}
\end{figure}

There are several important differences between an event camera and a human retina. The first and most obvious is that a retina has several stacked layers of processing (see Figure~\ref{fig:nowNextGoal}). The retina contains five types of neurons in stacked layers~\cite{purves2001neuroscience} to immediately process, organize, and distill localized information. Each retinal layer has various receptive field sizes to aid in processing prior to output via the optic nerve. Event cameras differ in that they \emph{process each pixel in isolation}. Independent processing has several major limitations including the event camera's inability to suppress noise. The retina increases the signal-to-noise ratio (SNR) by amplifying signal only~\cite{pahlberg2011visual}, while event cameras have no built-in noise suppression. For example, in photon-starved environments, event camera output can be dominated by noise. This results in large data rates, longer latency, and can make processing extremely challenging for the majority of applications. Recent research in this area has enabled on-chip processing that specifically mimics the nervous system within the retina. The combination of imaging and processing is called a pixel processor array (PPA). Perhaps one of the most sensible uses for PPAs is in data reduction via noise removal. By transitioning denoising algorithms upstream, the amount of data required for readout and transmission is significantly reduced as shown in Figure~\ref{fig:nowNextGoal}. This reduction in data would minimize downstream processing requirements, reduce latency, increase low-light applications, and compress bandwidth. The goal is to design a micro-network that could be utilized for SIMD processors like SCAMP~\cite{carey2013100,bose2019camera}. In fact, on-chip CNN inference is already possible for small recognition tasks~\cite{bose2020fully,bosehigh}. Regrettably, PPAs are still extremely limited in availability. For this reason, applications should focus on developing algorithms that could easily transition to PPA-like architectures once they become more widely available.

A second major difference is that event cameras encode a single feature of the scene (i.e. log-intensity change). In contrast, neurons in the retina are tuned to multiple unique ``features'' of the scene, including color, size, motion direction, and motion speed~\cite{nelson2007visual}. This lack of feature variability is a fundamental limitation to event cameras serving as a universal imaging platform, but it allows event cameras to achieve superior performance in specific scenarios and applications.

Finally, the retina exports information via parallel optic nerve connections to various parts of the brain. Event cameras, on the other hand, output all events to a single data stream that adds latency and is vulnerable to saturation during periods of large motion, high noise, and global intensity changes---an issue that continues to grow as event camera resolution and sensitivity increase. This large variance in global event rate is a significant issue in real-time and edge computing.

Figure~\ref{fig:nowNextGoal} shows the proposed system design aimed at mimicking naturally-occurring processes within neurons and moves closer to an end-to-end biologically-derived architecture. Ideally, a virtual retina would be implemented via an asynchronous architecture to mimic a human retina, but even using a synchronous approach would allow additional feature extraction and noise reduction. Rather than converting events into another format for synchronous processing, TORE volumes simply retain a history of recent asynchronous events within a neighborhood around each pixel. TORE volumes also allow for easy adaption of receptive field sizes and memory requirements.

\section{TORE Volumes}\label{sec:tore}

While event cameras are bio-inspired in their design, the representations used to group events have not remained true to this goal. Some representations convert the data to ``dense'' tensors that are surprisingly very sparse. Other methods apply temporal windowing that increases data size, destroys temporal information, and requires optimization based on scene motion and camera settings. TORE volumes are designed to leverage the fact that events are inherently a sparse representation. Additionally, TORE volumes require minimal, if any, tuning to achieve optimal performance.

\begin{figure}[htbp]
    \centerline{\includegraphics[width=1.0\linewidth]{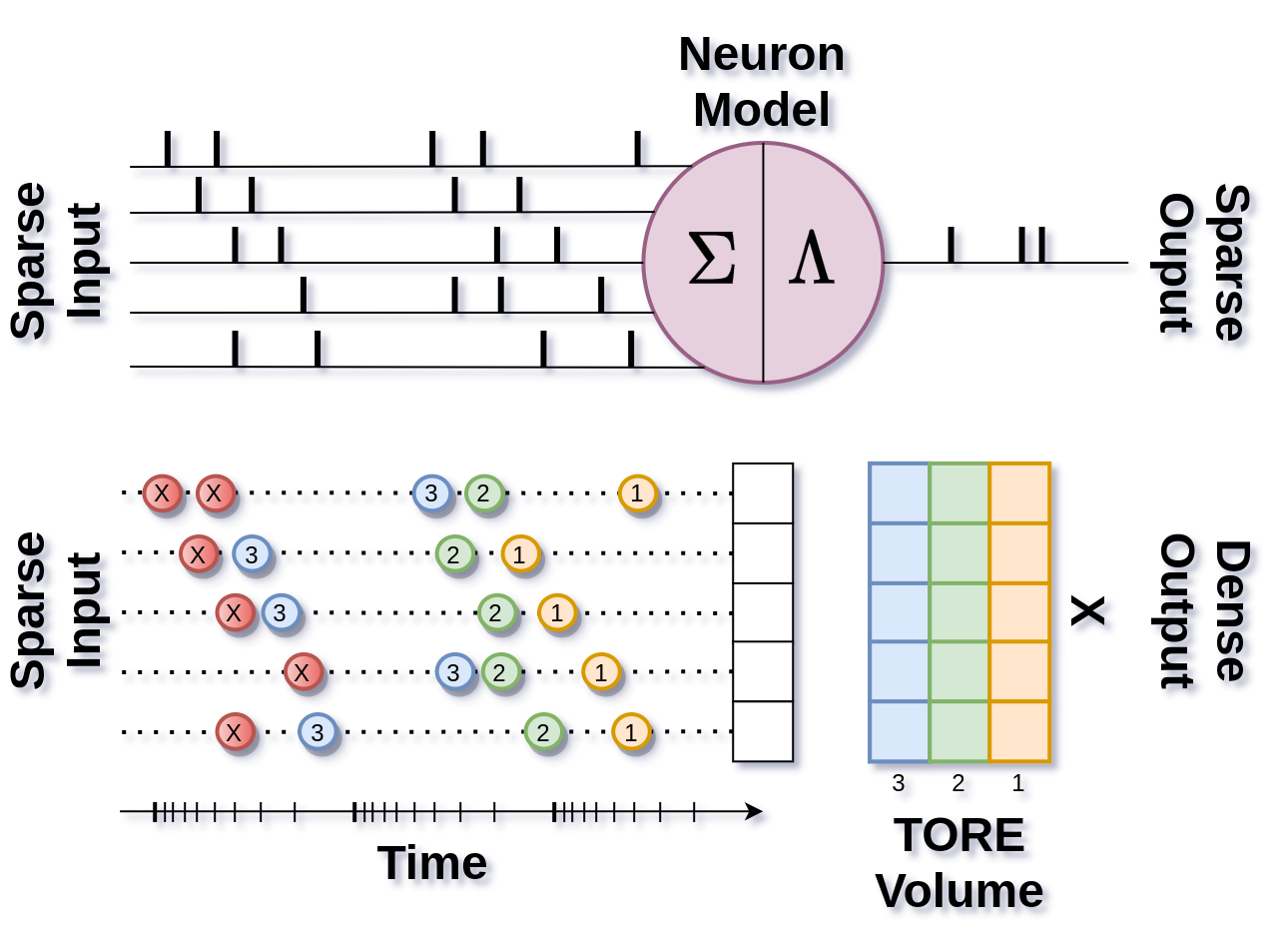}}
    \caption{2D representation of TORE volume generation. (Top) A simple neuron model takes multiple sparse inputs and generates one or more sparse output signals. The membrane potential (i.e. ``state'') of the neuron is impacted most by recent spikes in most LIF models. (Bottom) TORE volumes use a FIFO buffer to retain the most recent events at each location. Events beyond the buffer $k$ (shown here as $K=3$) are forgotten. A convolution on the TORE volume has the ability to approximate the output of a single neuron.}
    \label{fig:toreRepresentation}
\end{figure}

In the retina, neurons are connected to a localized set of detectors (i.e. rods and cones). The neurons maintain a continuous internal signal called a membrane potential. The membrane potential is often modeled as an exponentially decaying integrator with a reset threshold (i.e. Leaky-Integrate-and-Fire or LIF). Figure~\ref{fig:toreRepresentation} shows a simple neuron model driven by spikes from multiple inputs. To reconstruct this continuous signal accurately, you would need all prior spikes received by the neuron; however, due to the exponential decay of function, the signal is predominantly impacted by recent events. Therefore, if you were limited to fixed number of events as input to an estimation algorithm, it would make sense to prioritize the most recent events as these allow for the closest reconstruction of the continuous signal. This is exactly what TORE volumes accomplish by using a ``First-In, First-Out'' (FIFO) buffer at each pixel location. 

By moving representations from an arbitrary description to one based on biological systems, it is possible to improve accuracy across a wide range of applications. TORE volumes utilize the fact that data is already in a sparse format. Additionally, data is inherently prioritized by the timestamp of each event. Meaning older events are generally less important than more recent events. TORE volumes, therefore, present the first convolutional layer the ability to reconstruct the original sparse input. This ``replay'' ability allows the network to learn and extract features in the data in a similar fashion to retinal neurons. 2D convolution kernels combined with TORE volumes contain the rawest and most crucial information for a local area. Additionally, although TORE volumes are a dense representation, they do not involve constructing frames or time-windowing. Moreover, TORE volumes fill the dense representation, unlike time-windowed approaches that generate a sparsely-filled dense representation (see Figure~\ref{fig:toreRepresentation}).

TORE volumes are implemented using a per pixel polarity-specific FIFO queue $FIFO(x,y,p,k)$. The queue depth $K$ (i.e.~$k\in\{1,\dots,K\}$) is held constant as the addition of a new event $\bm{e}=(e_x,e_y,e_t,e_p)$ to the queue results in the removal of the oldest event, in the following manner:
\begin{align}\label{eq:fifo}
\begin{split}
    FIFO(e_x,e_y,e_p,K)&=FIFO(e_x,e_y,e_p,K-1)\\
    FIFO(e_x,e_y,e_p,K-1)&=FIFO(e_x,e_y,e_p,K-2)\\
    &\vdots\\
    FIFO(e_x,e_y,e_p,2)&=FIFO(e_x,e_y,e_p,1)\\
    FIFO(e_x,e_y,e_p,1)&=e_t.
\end{split}
\end{align}
Then, we define the TORE volume as the log-time difference between the current time $t$ and the $k$ most recent events stored in FIFO above, as follows:
\begin{align}\label{eq:tore1}
    TORE(x,y,p,k,t)=\log(t-FIFO(x,y,p,k)+1).
\end{align}

The logarithm in TORE promotes compression of large temporal differences, corresponding to events farther in past. This promotes hierarchy where higher significance is given to the more recent events. In some applications where it makes sense to impose a limit on the memory retention, a hard threshold is applied. Thus, we modify the definition of TORE in \eqref{eq:tore1} as:
\begin{align}\label{eq:tore2}
\begin{split}
    &TORE(x,y,p,k,t)=\\
    &\min(\log(t-FIFO(x,y,p,k)+1),\log(\tau)),
\end{split}
\end{align}
where $\tau$ is the maximum time. Time windowing is driven by scene content, but it is insensitive due to the log function in the equation. Experiments in this paper use $\tau=5\times 10^6$ (5 second) maximum memory retention.

It was also empirically shown that limiting the minimum time window of TORE volumes is beneficial. At the pixel level, sensor hardware has a limit for bandwidth (i.e. maximum firing rate), refractory period, and temporal resolution~\cite{gallego2019event}, introducing a minor uncertainty in event timestamps. Thus, the most recent event $e_t$ occurred close to current time $t$, temporal difference $t-FIFO(x,y,p,k)$ is dominated by such measurement noise and amplified by logarithm. To suppress this, we modify TORE volumes by also limiting minimum time sensitivity:
\begin{align}\label{eq:tore3}
\begin{split}
    &TORE(x,y,p,k,t)=\\
    &\max(\min(\log(t-FIFO(x,y,p,k)+1),\log(\tau)),\log(\tau')),
\end{split}
\end{align}
where $\tau'=150$ (150 microsecond) sensitivity was used for all experiments in this paper.


The approach presented above allows creating a TORE volume for any time period given a collection of events. Implementing TORE volumes as a real-time algorithm requires a buffer of size $K$ per pixel per polarity for FIFO, initialized to negative infinity. The computed TORE volume is a $2\times K\times H\times W$ or $2\times K$ layered $H\times W$ intermediary representation images, used by the subsequent processing steps (such as CNN). As events occur, the pixel-level FIFO is shifted asynchronously and the new timestamp is inserted in the first buffer position for that pixel. This structure is amenable to the throughput of high resolution DVS sensors.

The proposed TORE volumes are able to capture recent as well as past information about neuromorphic events without requiring temporal binning and windowing, event filtering, or obscuring polarity information. By design, TORE volumes are fast, memory efficient, and useful across a wide-range of applications. Whereas tuning parameters (including temporal window width) can easily affect performance accuracy for other algorithms, TORE volumes have minimal parameters (depth $K$ and temporal thresholds $\tau$ and $\tau'$) that are shown to be robust across multiple applications.

The benefit to the proposed approach is that the FIFO in \eqref{eq:fifo} can be updated asynchronously, while the TORE volumes in \eqref{eq:tore3} can be computed at time $t$ independent of the event generation timing. For instance, it is possible that the TORE volume generation at time $t$ is synchronized to the APS frame timing. The fact that TORE volumes lack the notion of a temporal window makes it robust to the speed of the objects represented in the scene. For example, it is possible to train using features that match the APS frame times, and perform inference at a much faster frame rates to simulate high-speed video, without any regard for temporal windowing. 

\section{Application: Event Denoising}\label{sec:denoise}

Event denoising is a critical application for event cameras. Under non-ideal conditions or camera settings, noise can dominate the available bandwidth of the sensor. This results in missed events, inaccurate event timestamps, reduced contrast sensitivity, and overall poor application performance. As discussed previously in Section~\ref{sec:intro_retina}, the human retina reduces visual stimulus noise while amplifying signal prior to transmission over the optic nerve. A similar denoising mechanism employed within event cameras would greatly benefit this technology. 

\begin{figure}[htbp]
    \centerline{\includegraphics[width=1.0\linewidth]{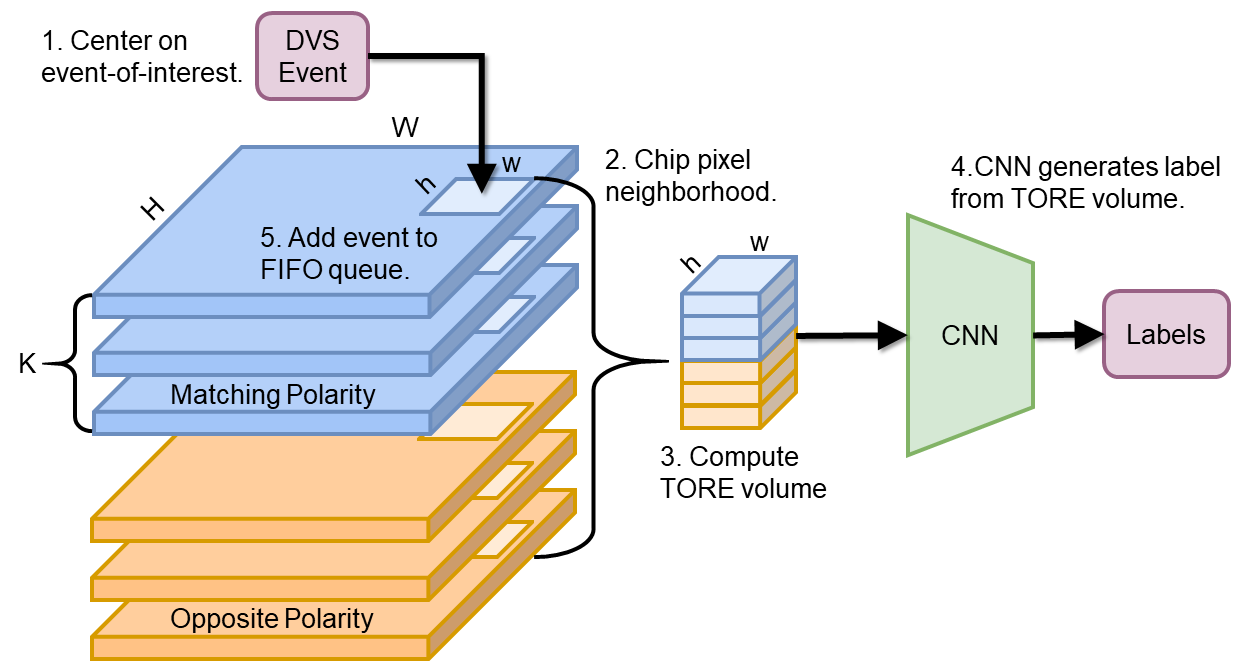}}
    \caption{TORE volume is generated from $K$ most recent events in $m\times m$ neighborhood of event-of-interest. All surfaces are concatenated and passed to classification network. CNN performs binary classification to yield a denoising label.}
\label{fig:toreChip}
\end{figure}

\begin{figure*}[tb]
\centering
\begin{tabular}{m{0.01\textwidth}m{0.28\textwidth}m{0.28\textwidth}m{0.28\textwidth}}
\rotatebox{90}{Noisy (Raw)} & \includegraphics[width=1.05\linewidth]{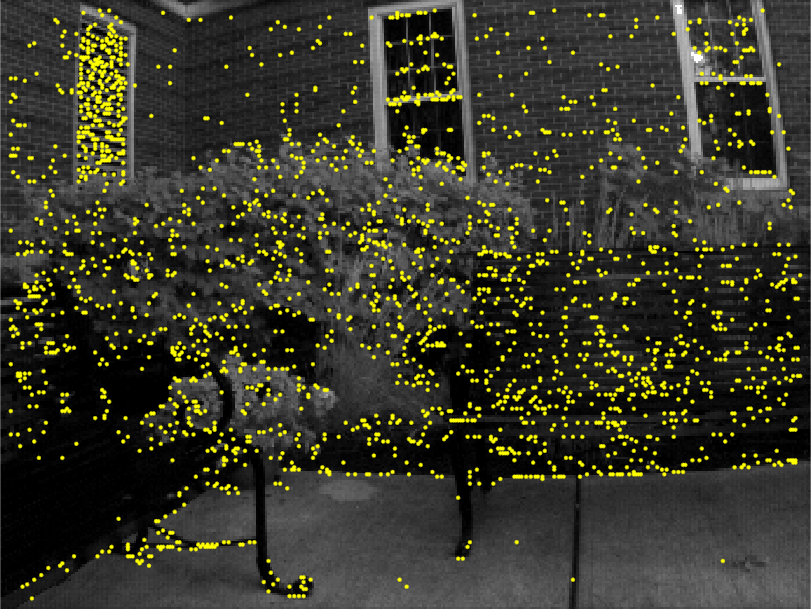} & \includegraphics[width=1.05\linewidth]{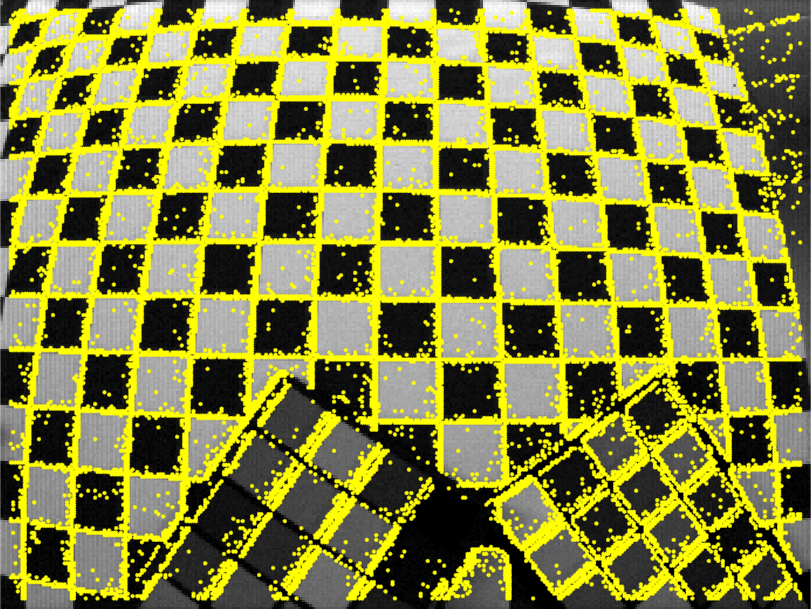} & \includegraphics[width=1.05\linewidth]{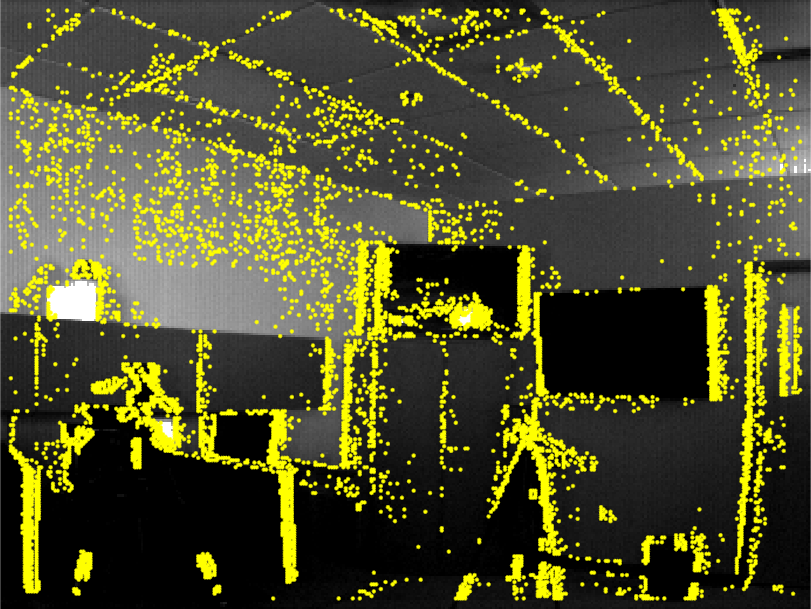}\\
\rotatebox{90}{BAF~\cite{delbruck2008frame}} & \includegraphics[width=1.05\linewidth]{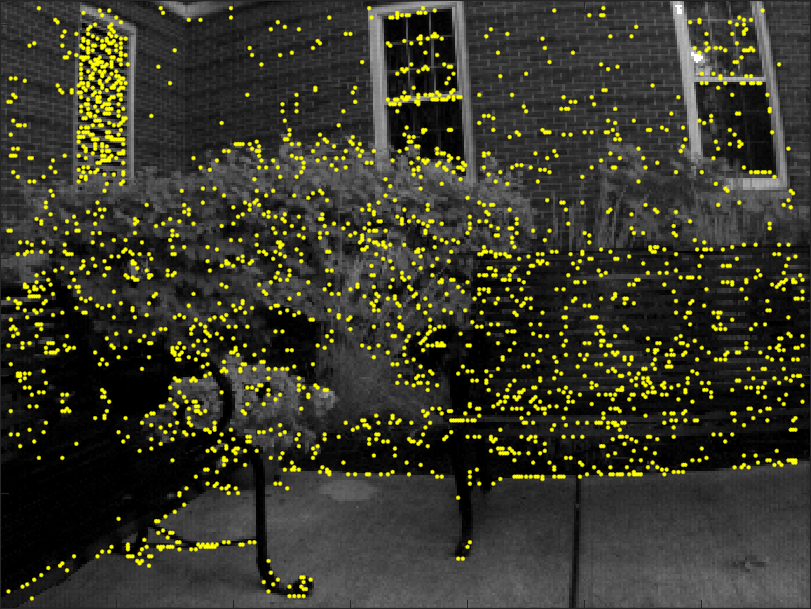} & \includegraphics[width=1.05\linewidth]{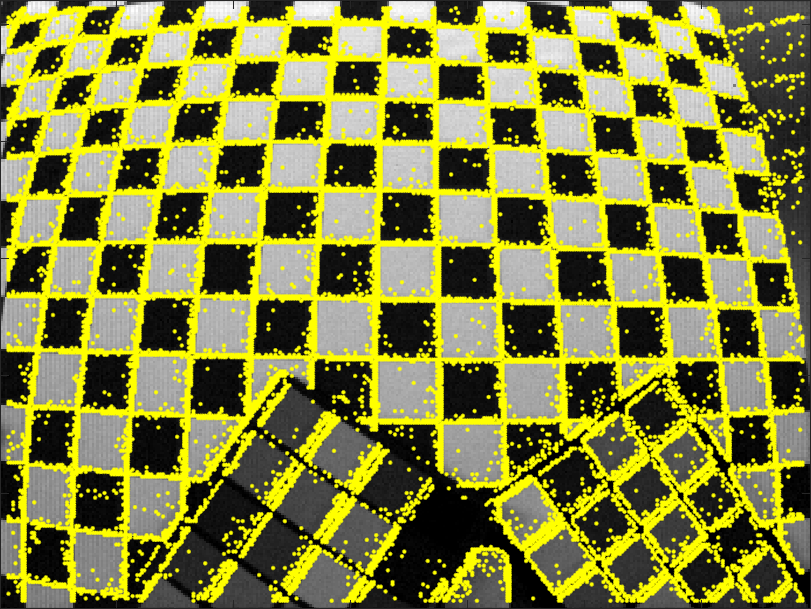} &
\includegraphics[width=1.05\linewidth]{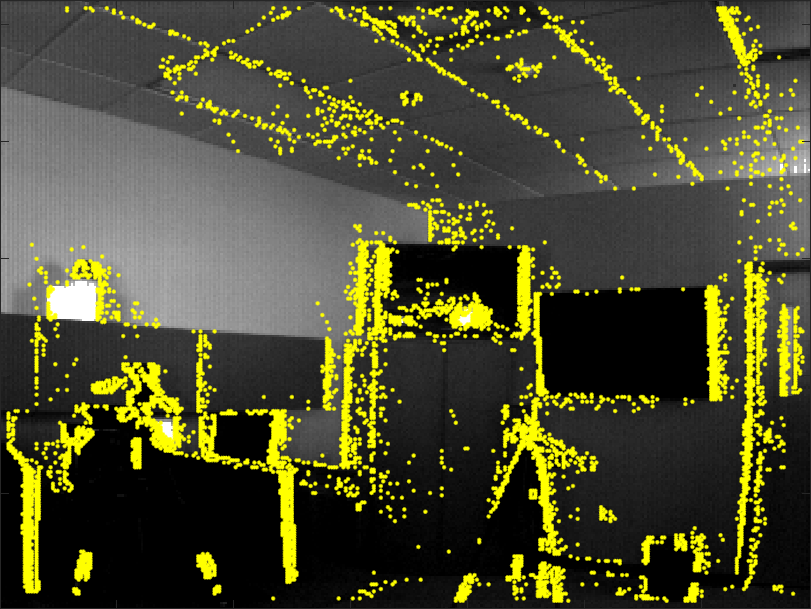} \\
\rotatebox{90}{\textbf{TORE (proposed)}} & \includegraphics[width=1.05\linewidth]{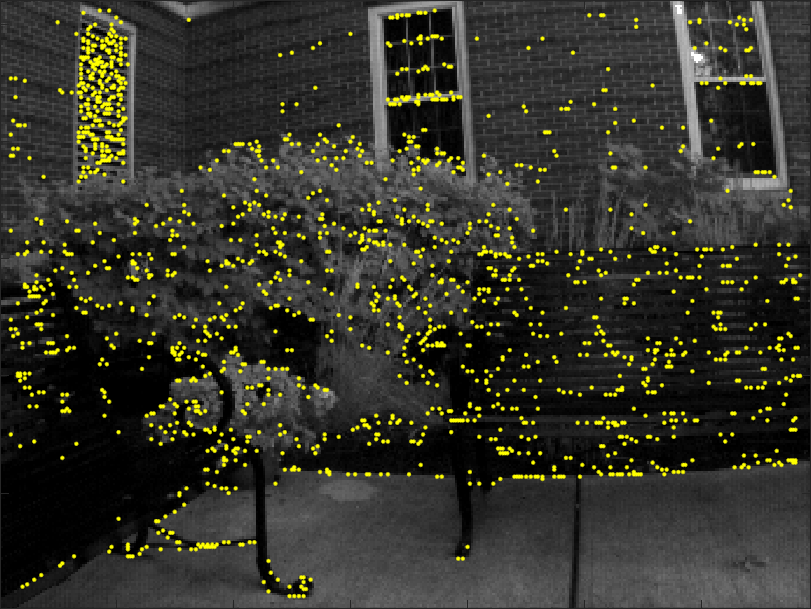} & \includegraphics[width=1.05\linewidth]{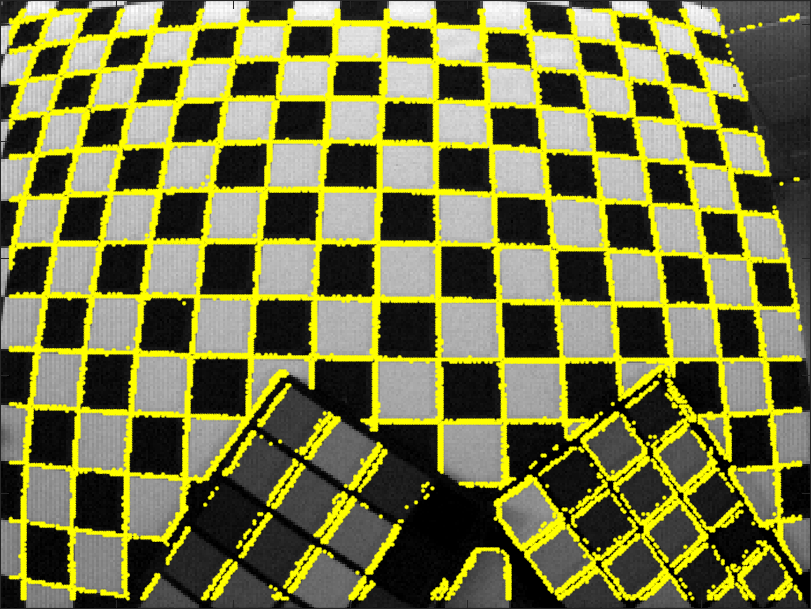} &
\includegraphics[width=1.05\linewidth]{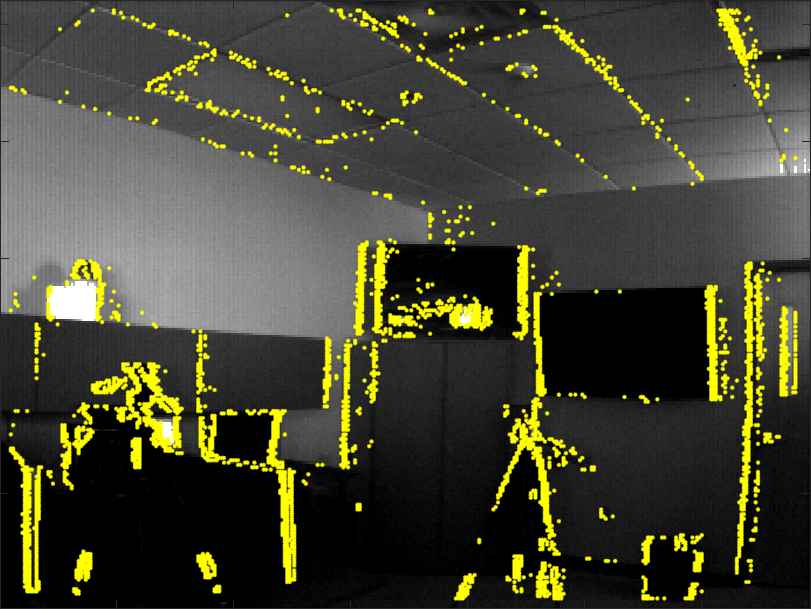} \\
\multicolumn{1}{r}{} & \multicolumn{1}{c}{Benches} & \multicolumn{1}{c}{BigChecker} & \multicolumn{1}{c}{LabFast}
\end{tabular}%
\caption{Event denoising results from DVSNOISE20 dataset overlaid on corresponding APS image. (First Row) All events generated by the event camera. (Second Row) Denoised events using BAF method.  (Third Row) Denoised events using proposed method. The proposed denoising algorithm does not use APS image information, but most events not corresponding to edges visible in the APS image have been removed as likely noise.}
\label{fig:noiseImages}
\end{figure*}

Prior works in event noise suppression relied largely on handcrafted features. Owing to the fact that the exact noise distribution of DVS sensors is unknown, these existing approaches are adhoc or based on empirical evidences. We, however, approach this problem from a machine learning perspective, focused on developing a spatially localized model using training data containing real sensor noise.
Specifically, we used a simple three layer CNN with each layer composed of a convolution, batch normalization, relu, and max pooling layer. After the third layer, a dropout layer and two fully connected layers completed the network. The input was a TORE volume using $K=7$ and patches of size $9\times 9$ centered around an event-of-interest. The output of this network is a binary classifier, indicating whether the event-of-interest is noise or signal, trained with a dataset we developed called DVSNOISE20~\cite{baldwin2020event}. DVSNOISE20 is one of the largest event camera datasets available containing raw event data, APS frames, and IMU measurements, and it provides per pixel labels for noise vs.~real events derived from APS and IMU data. The DVSNOISE20 dataset contains a combination of 16 indoor and outdoor scenes of noisy, real-world data with three recordings per scene. To avoid the possibility of learning specific scene content, the dataset was split into training/testing by scene. For this research, 13 scenes were used for training and three scenes were used for testing. Scenes were partitioned to obtain a test sample of indoor/outdoor, slow/fast motion, and complex/simple scenes. As outlined in prior work, training and benchmarking were restricted to events occurring during APS exposure time since labeling is only valid during this time. Table~\ref{tab:denoise} shows how spatially localized TORE volume patches with deep FIFO structure can improve state-of-the-art for event denoising. This is a very promising conclusion and opens the door for further work to move event denoising to an architecture like PPAs (discussed in Section~\ref{sec:intro_retina}). As evidenced by the qualitative evaluation of denoising results in Figure~\ref{fig:noiseImages}, spatially localized TORE volume patches can yield convincing denoising performance.

\begin{table}
\centering
\caption{RPMD scores for event denoising on DVSNOISE20 (sampled) dataset. TORE volume obtains higher accuracy than EDnCNN while reducing spatial area and representation size.}
\label{tab:denoise}
\scalebox{0.98}{
\begin{tabular}{cccccc} 
\toprule
    Denoising & Input & \multicolumn{3}{c}{Scene (RPMD)}  &          \\ 
\cline{3-5}
Method & Size & Benches & BigChecker & LabFast & Mean \\ 
\hline\hline
Noisy (Raw)                              & ---     & 150 & 344 & 172 & 222.0\\
FSAE~\cite{mueggler2017fast}             & 1x1     & 160 & 284 & 164 & 202.7\\
IE~\cite{baldwin2019inceptive}           & 1x1     & 156 & 241 & 126 & 174.3\\
IE+TE~\cite{baldwin2019inceptive}        & 1x1     & 100 & 75  & 58  & 77.7\\
BAF~\cite{delbruck2008frame}             & 3x3     & 44  & 110 & 99  & 84.3\\
EDnCNN~\cite{baldwin2020event}           & 25x25x4 & 30  & 59  & 61  & 50.0\\
\bf{TORE (ours)}                         & 9x9x14  & 18  & 45  & 67  & \textbf{43.6}\\
\bottomrule
\end{tabular}}
\end{table}

\section{Application: Frame Reconstruction}\label{sec:reconstruction}

Intensity frame reconstruction (as an image or video sequence) is another canonical task for event cameras. Intensity frames can be reconstructed at high frame rate yielding high-speed video capability with very low data rates. The intensity frames reconstructed from DVS data enjoy a considerably larger dynamic range compared to APS, yielding evenly exposed images in shadows and highlights (note that DVS readout circuits themselves lack the notion of exposure). In addition, intensity frames reconstructed from DVS exhibit sharp edges, unlike the APS images that suffer from motion blur when the camera or the scene motion is fast relative to exposure time.

There are at least two major challenges to achieving high-quality reconstruction. First, event cameras currently have significantly lower contrast sensitivity than standard imaging sensors. This can result in the loss of fine details and edges during the reconstruction process. Second, the DC component (average intensity) of the underlying intensity signal is never measured. This is typically overcome by min-max contrast stretching applied to the reconstructed frames. While contrast stretch improves the appearance of the frames, such enhancement technique is not photometrically faithful and can dominate loss functions (i.e. MSE and SSIM). Moreover, minimum and maximum values are sensitive to noise.

In this work, intensity frame reconstruction was achieved by combining $K=4$ depth TORE volumes computed for the entire sensor focal plane and a U-Net architecture~\cite{ronneberger2015u} with five encoder/decoder levels. The network was trained to reconstruct histogram equalized APS frames in DVSNOISE20 dataset by using TORE volumes computed from the corresponding event data. Prior work has also used histogram equalization for target images. This is because the DC value at each pixel is not reflected in the event data obtained from DVS circuit. Without DC subtraction or histogram equalization, the loss function is dominated by DC errors. 

The proposed U-Net was trained by synchronizing intensity frame reconstruction output with APS frames. Although impractical for real-time, a network trained in such a manner can be used to generate frames at a very high rate without retraining, due to the lack of temporal windowing in TORE volumes. This unique and valuable feature enables a single well-trained network to function across a wide-range of use cases. An image reconstruction network trained on the DVSNOISE20 and HQF datasets can reconstruct video from a fast moving drone-mounted event camera, for example. The key to this technique is to train using APS frames free of blurring, saturation, and underexposure---true of both datasets considered in this work~\cite{baldwin2020event,stoffregen2020reducing}. 

\begin{figure}[htbp]
\centering
\begin{tabular}{m{0.01\linewidth}m{0.9\linewidth}}
\rotatebox{90}{bigChecker} & \includegraphics[width=1\linewidth]{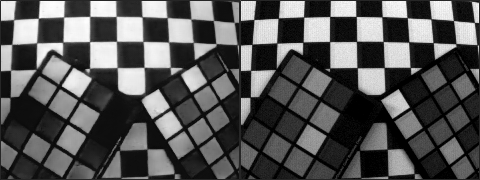}\\
\rotatebox{90}{conference} & \includegraphics[width=1\linewidth]{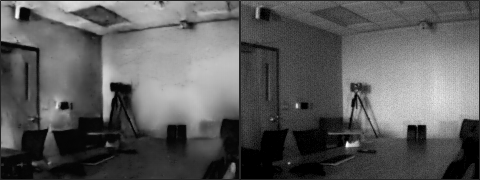}\\
\rotatebox{90}{calibration} & \includegraphics[width=1\linewidth]{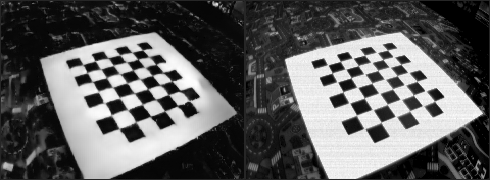}\\
\rotatebox{90}{poster\_6dof} & \includegraphics[width=1\linewidth]{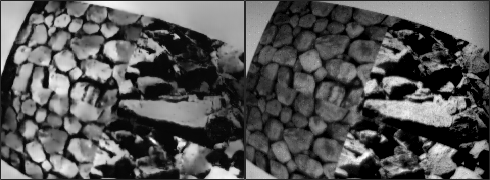}\\
\rotatebox{90}{slider\_depth} & \includegraphics[width=1\linewidth]{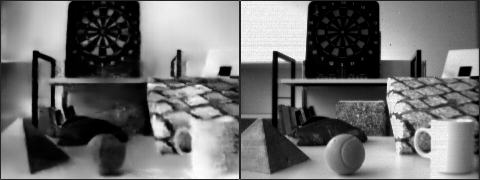}\\
\rotatebox{90}{dynamic\_6dof} & \includegraphics[width=1\linewidth]{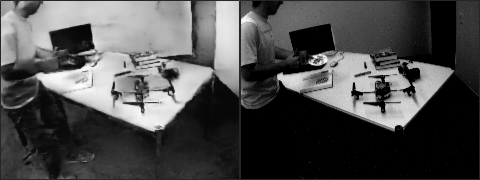}\\
\multicolumn{1}{r}{} & \multicolumn{1}{l}{\hspace{0.95cm} Reconstruction \hspace{1.8cm} Ground Truth}\\
\end{tabular}
\caption{Example of image reconstruction using event representations. The left image was reconstructed using only event representations and the right image was captured from the APS output (i.e. truth). The first two rows represent the test scenes from the DVSNOISE20 dataset (i.e. trained on similar conditions to the test data). The bottom four rows show reconstruction from the ECD dataset. Animations are available via the project website.}
\label{fig:imReconDVSNOISE20}
\end{figure}


Figure~\ref{fig:imReconDVSNOISE20} shows example reconstructions from the test sequences in DVSNOISE20 and ECD datasets. Inference on TORE-based U-Net runs at over 350fps on a single Nvidia GeForce RTX 2080 Ti. In Table \ref{tab:imrecon}, TORE volume event reconstruction was compared to the results from state-of-the-art E2VID method using the Event Camera Dataset (ECD)~\cite{mueggler2017event}. TORE volumes match E2VID performance using MSE but had lower SSIM score. A contributing factor to this is the mismatch between the APS camera settings in DVSNOISE20 dataset (that TORE-based U-Net was trained on) and ECD (that we tested on), introducing differences in image appearances. E2VID also imposes temporal consistency by incorporating past frames in the reconstruction of the present frame. Incorporating this strategy into TORE-based U-Net would likely improve the MSE and SSIM scores further. Regardless of the method, while image reconstruction from event cameras has undergone meaningful growth in recent years, there is still significant improvements needed in resolution, sensitivity, and noise reduction before event cameras can bridge the current gap in replicating high-speed videos.

\begin{table}
\centering
\caption{Image reconstruction scores for EV2VID and TORE on the Event Camera Dataset. Overall, EV2VID outperforms U-Net/TORE reconstruction, but the simple architecture of U-Net/TORE can achieve quality images and outperform EV2VID in some scenes.}
\label{tab:imrecon}
\scalebox{1.0}{
\begin{tabular}{cccccc} 
\toprule
     & \multicolumn{2}{c}{(MSE)} & & \multicolumn{2}{c}{(SSIM)}            \\ 
\cline{2-3} \cline{5-6}
Dataset & EV2VID & TORE & & EV2VID & TORE\\ 
\hline\hline
boxes\_6dof & \textbf{0.04} & 0.05 & & \textbf{0.63} & 0.53\\
calibration & 0.04 & \textbf{0.03} & & \textbf{0.52} & \textbf{0.52}\\
dynamic\_6dof & 0.08 & \textbf{0.07} & & 0.50 & \textbf{0.55}\\
office\_zigzag & \textbf{0.05} & \textbf{0.05} & & \textbf{0.50} & 0.47\\
poster\_6dof & \textbf{0.04} & \textbf{0.04} & & \textbf{0.68} & 0.59\\
shapes\_6dof & 0.10 & \textbf{0.06} & & 0.44 & \textbf{0.63}\\
slider\_depth & 0.06 & \textbf{0.05} & & \textbf{0.61} & 0.59\\
\bottomrule
mean & 0.06 & \textbf{0.05} & & \textbf{0.56} & 0.55\\
\bottomrule
\end{tabular}}
\end{table}

\begin{figure*}[tb]
    \centering
    \includegraphics[width=1\linewidth]{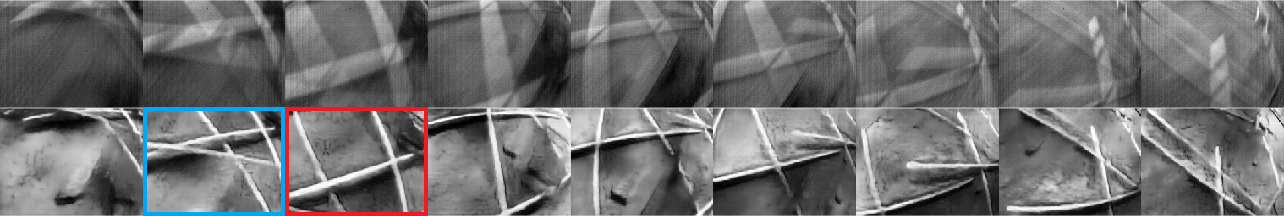}
   \caption{A sequence of nine APS frames from the UZH-FPV Drone Racing dataset (top). This dataset is extremely dissimilar to the training data, but reconstruction from DVS-only data removes blurring caused by the extreme camera motion (bottom).}
\label{fig:UAVrecon}
\end{figure*}

\begin{figure*}[tb]
    \centering
    \includegraphics[width=1\linewidth]{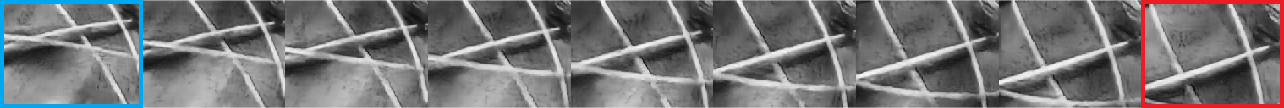}
    \caption{A sequence of DVS-only reconstructed frames from the UZH-FPV Drone Racing dataset. An additional seven frames are generated from TORE representations between the APS frames to create a video with a frame rate $8\times$ the APS sensor.}
\label{fig:fastUAV}
\end{figure*}

Frame reconstruction can be extended to generate video at a very fast temporal resolution. The UZH-FPV Drone Racing dataset~\cite{Delmerico19icra} contains data from a DVS camera mounted onto a racing drone. APS images from the camera are blurry due to the extremely fast camera motion. Using only TORE representations from DVS events, it is possible to generate frames with significantly less artifacts since the TORE-based U-Net was trained on non-blurry APS images. Figure~\ref{fig:UAVrecon} shows a sequence of nine consecutive frames from the \emph{indoor\_45\_11\_davis} recording. Reconstructed frames are sharper and provide an improved visualization of the scene. Additionally, since the network design does not use temporal windowing, it is easy to adjust the output frame rate and generate images faster than the APS sensor. Figure~\ref{fig:fastUAV} shows frames generated at $8\times$ the rate of the APS sensor. TORE-based U-Net provides smooth motion flow as the images are based solely on DVS events with high temporal resolution.

\section{Application: Classification}\label{sec:classification}

TORE volumes are ideal for classification due to their low complexity and ability to encode fine details. To demonstrate comprehensive utility, TORE volumes were used as input to a CNN-based classifier for several challenging datasets. For all datasets evaluated using TORE volumes, no noise or bad pixel removal was preformed prior to representation generation, as we found empirically that combinations of TORE volumes and CNNs were generally robust to noise and pixel artifacts. 
Additionally, TORE volumes are causal and, therefore, do not introduce latency. Due to the lack of temporal windowing, TORE-based networks can be evaluated at any rate without retraining. We contrast this to methods that incorporate temporal windowing, where there is a delay in the processing of the events occurring early in the window. 

Following the examples of existing event-based state-of-the-art classification work~\cite{baldwin2019inceptive,rebecq2019events}, classification in this work was carried out by a CNN using a pre-trained network (GoogLeNet\cite{szegedy2015going}) with transfer learning. Prior to training, TORE volumes are resized per channel to a size of $224\times224$. The TORE volume depth was three per polarity, making the entire representation $224\times224\times6$. To account for more than three channels (RGB of original GoogLeNet), the network was recreated as a 3D convolutional input network and the pre-trained weights and biases were replicated for the two polarities. This allowed GoogLeNet to run per polarity with single set of shared weights. Just prior to the fully connected layers, the tensors are reshaped and concatenated to form a vector twice the normal length in GoogLeNet. This restructuring from 2D to 3D allows the use of deep networks pre-trained on datasets significantly larger than those available in the event camera community. All fully connected layers were modified to have bias and weight learn rates of $10\times$. Although there has been progresses made since the introduction of GoogLeNet, we found it to work well on event representations and to have a good balance of total parameters, size, and accuracy. The impact of using a different pre-trained network was not exhaustively tested. There are likely networks that execute faster or generate higher accuracy for various classification tasks.


In this study, we performed classification on a number of event camera datasets, and compared the accuracy of TORE volume-based classification to several state-of-the-art methods  \cite{baldwin2019inceptive,bi2019graph,rebecq2019events,amir2017low,shrestha2018slayer,vasudevan2020introduction,gehrig2019end}. Unless otherwise specified, the experiments followed the procedures of the prior work to partition the dataset into training/testing samples. The results are summarized in Table~\ref{tab:objaccuracy}. TORE volumes have achieved state-of-the-art performance on five of the six tested datasets.

N-CARS~\cite{sironi2018hats} is a large public dataset for car classification. This real-world dataset is composed of 12,336 car samples and 11,693 background samples. The event camera was mounted behind the windshield of a moving car during collection, and each sample contains exactly 100 milliseconds of data with 500 to 59,249 events per sample. The number of samples and variety for each class makes this dataset ideal for machine learning, yet having only two classes reduces the challenge of this dataset. TORE volume-based classification beat the state-of-the-art scores for this dataset.

The N-MNIST dataset~\cite{orchard2015converting} is an event camera version of MNIST~\cite{lecun1998gradient}. The dataset was created by moving an event camera that was focused on an LCD monitor displaying the original MNIST data. The dataset is broken down into 60k training and 10k test sequences. As with MNIST, there are 10 total classes, one for each digit 0--9. Classification using TORE volumes achieve the highest accuracy on this dataset to date.

N-Caltech101~\cite{orchard2015converting} is an event camera version of the Caltech101~\cite{fei2004learning} dataset. It was created using the same methods as N-MNIST. The dataset consists of 100 object classes plus a background class. This is a challenging dataset based on the large number of classes as well as the unbalanced number of samples within each class. Because this dataset does not contain a specified train/test split, existing studies have randomly partitioned training and testing sets, making it difficult to compare the accuracy of one method to the accuracy of another. Nevertheless, we followed the example of the original paper in \cite{orchard2015converting}---testing using 15 random samples from each class---and achieved the state-of-the-art accuracy of $79.8$\%. HATS-ResNet18, E2VID, and EST reported accuracy using a random percentage split, and their accuracy scores are marked with * in Table \ref{tab:objaccuracy}. When we tested TORE volume-based classification in the same manner, our accuracy improved to $83.4$\%, which is second highest among scores marked with $*$. Readers are encouraged to interpret these scores with caution because the different instantiations of random train/test assignment influence the accuracy of unbalanced datasets.


The DVS128 Gesture~\cite{amir2017low} dataset contains continuous recordings of 11 different gestures. SNN-based classification methods in \cite{shrestha2018slayer,amir2017low} output continuously and report the accuracy after applying a temporal smoothing filter. Current state-of-the-art on this dataset is a PointNet++ design~\cite{wang2019space}, which is a CNN designed to directly process point cloud data, outputting results every 25ms, and applying temporal averaging filter of 225ms. We followed the same configuration---TORE volume-based classification algorithm outputs gesture classification every 25ms, to which we applied temporal averaging filter of 225ms. We achieved classification accuracy of $96.16$\%, higher than the PointNet++ design.


SL-ANIMALS-DVS~\cite{vasudevan2020introduction} is a collection of 59 seated subjects performing different Spanish sign language gestures for 19 different animals. The dataset contains four different collection environments (indoor, sunlight, imse, and dc). This dataset has been used to evaluate the classification accuracy of SNN-based methods in \cite{wu2018spatio,shrestha2018slayer} based on their filtered continuous classification results. To the best of our knowledge, no synchronous/CNN based classifications have been reported. For this reason, we follow the evaluation strategy of PointNet++ design described above~\cite{wang2019space} (25ms output rate, 225ms temporal averaging)to evaluate TORE classification accuracy. We achieved the highest classification accuracy of $85.05$\%.


Finally, ASL-DVS~\cite{bi2019graph} consists of 24 classes of hand recordings where each class represents a different letter from the American Sign Language (ASL) set. There are over 100k total samples of 100ms each, making this one of the largest event camera datasets. For evaluation, we matched prior work and performed a random split of 20\% for testing and 80\% for training. We outperformed the prior method in accuracy with nearly perfect accuracy scores.


\begin{table*}[htbp]
  \centering
  \caption{Classification accuracy compared to state-of-the-art methods on several public datasets. Number of classes for each dataset is shown in parenthesis in the first row. Note: N-Caltech101 is a highly imbalanced dataset without an official train/test label. The original paper used a random sample of 15 images for testing from each class. Other papers (marked with *) use a percentage split on the dataset.}
    \setlength{\tabcolsep}{4pt}
    \begin{tabular}{lcccccccc}
    \toprule
    Algorithm                                   & Classifier & N-CARS(2) & N-MNIST(10) & N-Caltech101(101) & DVS128 Gesture(11) & SL-ANIMALS-DVS(19) & ASL-DVS(24)\\
    \midrule
    HOTS~\cite{lagorce2016hots}                 & SVM        & 0.624     & 0.808       & 0.210             & ---                & ---         & ---   \\
    HATS~\cite{sironi2018hats}                  & SVM        & 0.902     & 0.991       & 0.642             & ---                & ---         & ---   \\
    H-First~\cite{orchard2015hfirst}            & SNN        & 0.561     & 0.712       & 0.054             & ---                & ---         & ---   \\
    Gabor~\cite{lee2016training,neil2016phased} & SNN        & 0.789     & 0.837       & 0.196             & ---                & ---         & ---   \\
    STBP~\cite{wu2018spatio}                    & SNN        & ---       & ---         & ---               & ---                & 0.714       & ---   \\
    SLAYER~\cite{shrestha2018slayer}            & SNN        & ---       & 0.992       & ---               & 0.936              & 0.780       & ---   \\
    TN\_E1~\cite{amir2017low}                   & SNN        & ---       & ---         & ---               & 0.946              & ---         & ---   \\
    Event Clouds~\cite{wang2019space}           & PointNet++ & ---       & ---         & ---               & 0.953              & ---         & ---   \\
    HATS-ResNet18\cite{rebecq2019events}        & CNN        & 0.904     & ---         & 0.700*            & ---                & ---         & ---   \\
    E2VID~\cite{rebecq2019events}               & CNN        & 0.910     & 0.983       & \textbf{0.866}*   & ---                & ---         & ---   \\
    RG-CNN~\cite{bi2019graph}                   & CNN        & 0.914     & 0.990       & 0.657             & ---                & ---         & 0.901 \\
    EST~\cite{gehrig2019end}                    & CNN        & 0.925     & ---         & 0.817*            & ---                & ---         & ---   \\
    IETS~\cite{baldwin2019inceptive}            & CNN        & 0.976     & ---         & ---               & ---                & ---         & ---   \\
    TORE (ours)                                 & CNN & \textbf{0.977} & \textbf{0.994} & \textbf{0.798}/0.834* & \textbf{0.962} & \textbf{0.851} & \textbf{0.996}\\
    \bottomrule
    \end{tabular}%
  \label{tab:objaccuracy}%
\end{table*}%

\section{Application: Human Pose Estimation}\label{sec:pose}

Human pose estimation has several challenges, including self occlusions of joints, 2D to 3D pose reconstruction, and motion blur. For example, rapid motions (i.e. arm-waving, jumping, spinning, etc.) can cause image blurring in conventional cameras. Although some challenges can be offset by higher frame rates, this can lead to further issues such as data transfer and additional hardware requirements.

\begin{figure*}[htbp]
\centering
    \includegraphics[width=0.9\linewidth]{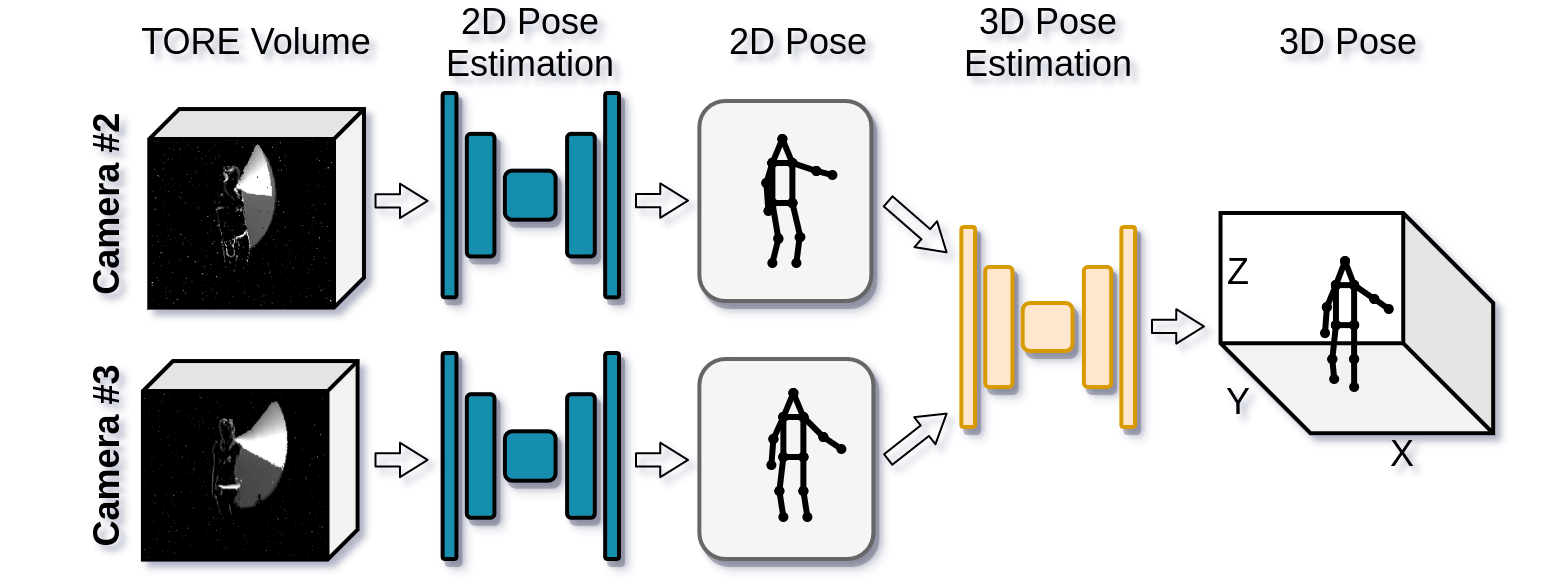}
   \caption{Overview of proposed pose estimation approach. An event representation for each camera is used as input to a small CNN (blue) to estimate a 2D pose for 13 joint positions. Joint positions from 1--4 cameras (two shown) are then passed to a second CNN (yellow) to estimate the overall 3D pose. The 2D pose estimation does not require a fixed frame rate and can output 2D pose at any temporal resolution once trained.}
\label{fig:poseOverview}
\end{figure*}

Event cameras cannot address all these issues, but they are well-suited to encode motions at rates sufficient to overcome the challenges associated with even the fastest human motion. TORE volumes are ideal for human pose estimation because they maintain fine temporal resolution along with a history of past events. We tested performance against the DHP19 dataset~\cite{calabrese2019dhp19}. In particular, we simply replaced the event representations with TORE volumes, while keeping identical CNN architecture in~\cite{calabrese2019dhp19}, and demonstrate an improved 2D pose estimation of 23\%. Additionally, we replaced the two camera triangulation in~\cite{calabrese2019dhp19} with a state-of-the-art 2D to 3D pose estimation network originally proposed for traditional cameras~\cite{liu2020attention} (see Figure~\ref{fig:poseOverview}). This further improves overall accuracy. In total, this design improves performance by 29\% over previous state-of-the-art methods.

\subsection{2D Pose Estimation}
\label{sec:2dpose}

\begin{figure}[htbp]
    \centering
    \includegraphics[width=1.0\linewidth]{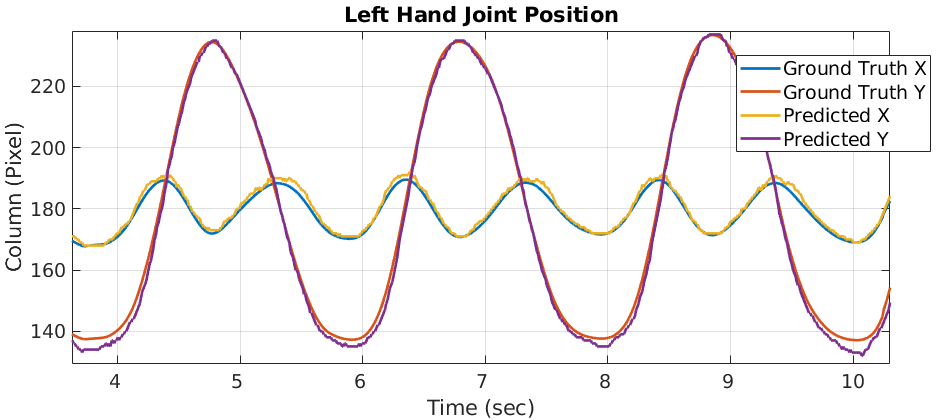}
   \caption{Track of left hand waving from camera \#3. Ground truth location, shown in blue/red, is the 2D position calculated from the motion tracking system projected into the sensor coordinates. The network estimated location, shown in yellow/purple, is estimated using only sparse event data.}
\label{fig:leftHandTrack}
\end{figure}

\begin{figure}[htbp]
\centering
\begin{tabular}{m{0.01\linewidth}m{0.4\linewidth}m{0.4\linewidth}}
\rotatebox{90}{Example 1} & \includegraphics[width=1.07\linewidth]{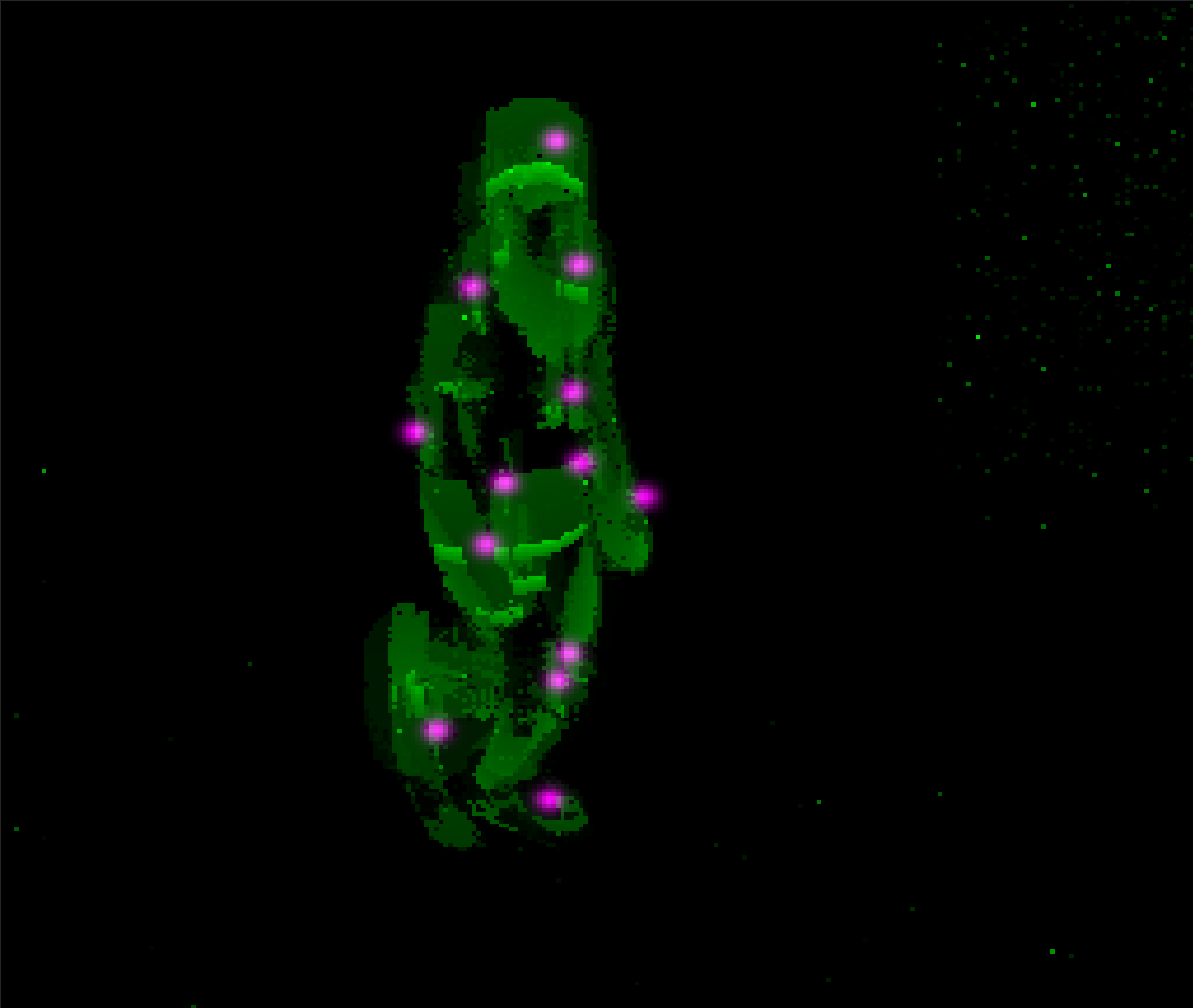} & \includegraphics[width=1.07\linewidth]{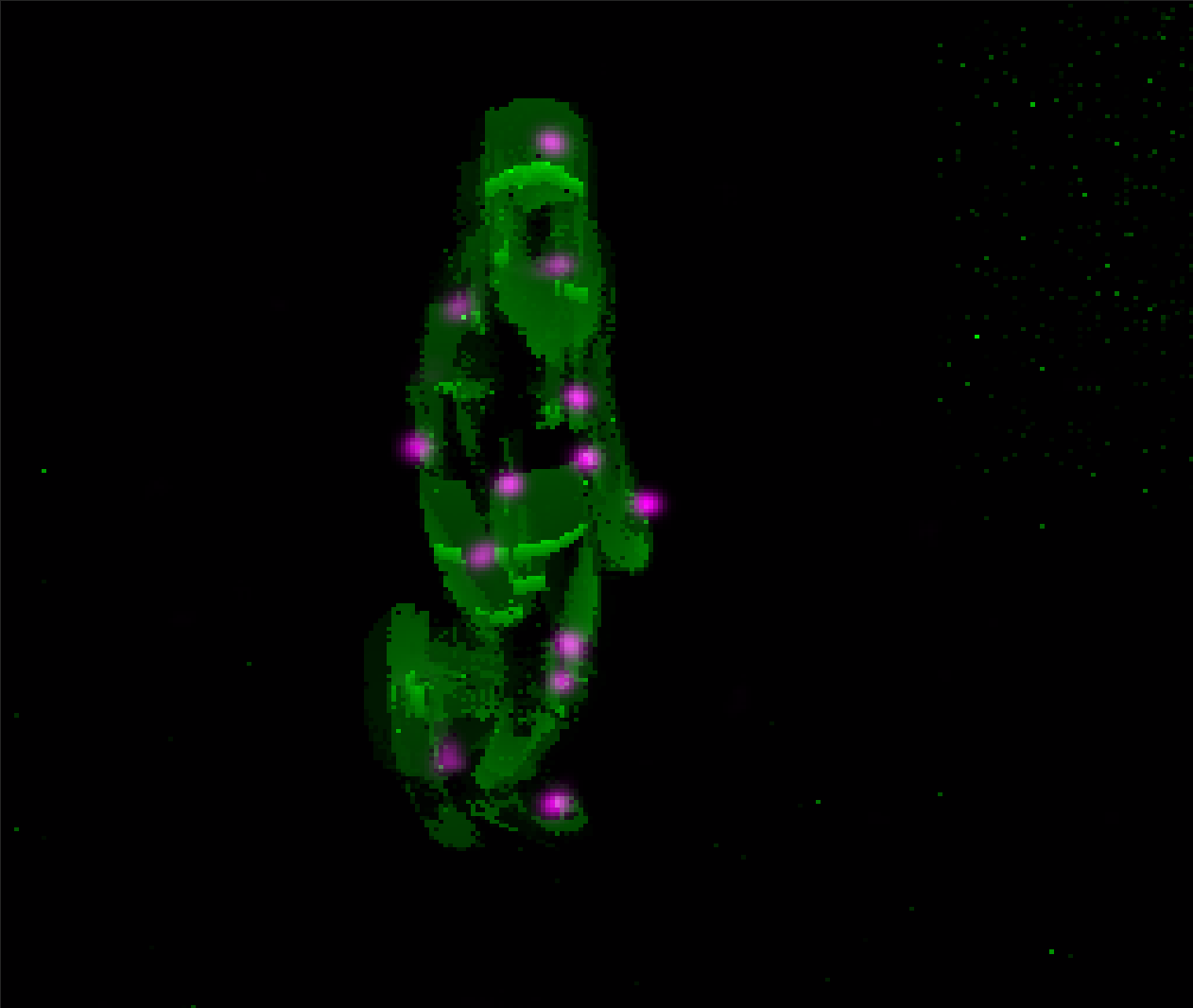}\\
\rotatebox{90}{Example 2} & \includegraphics[width=1.07\linewidth]{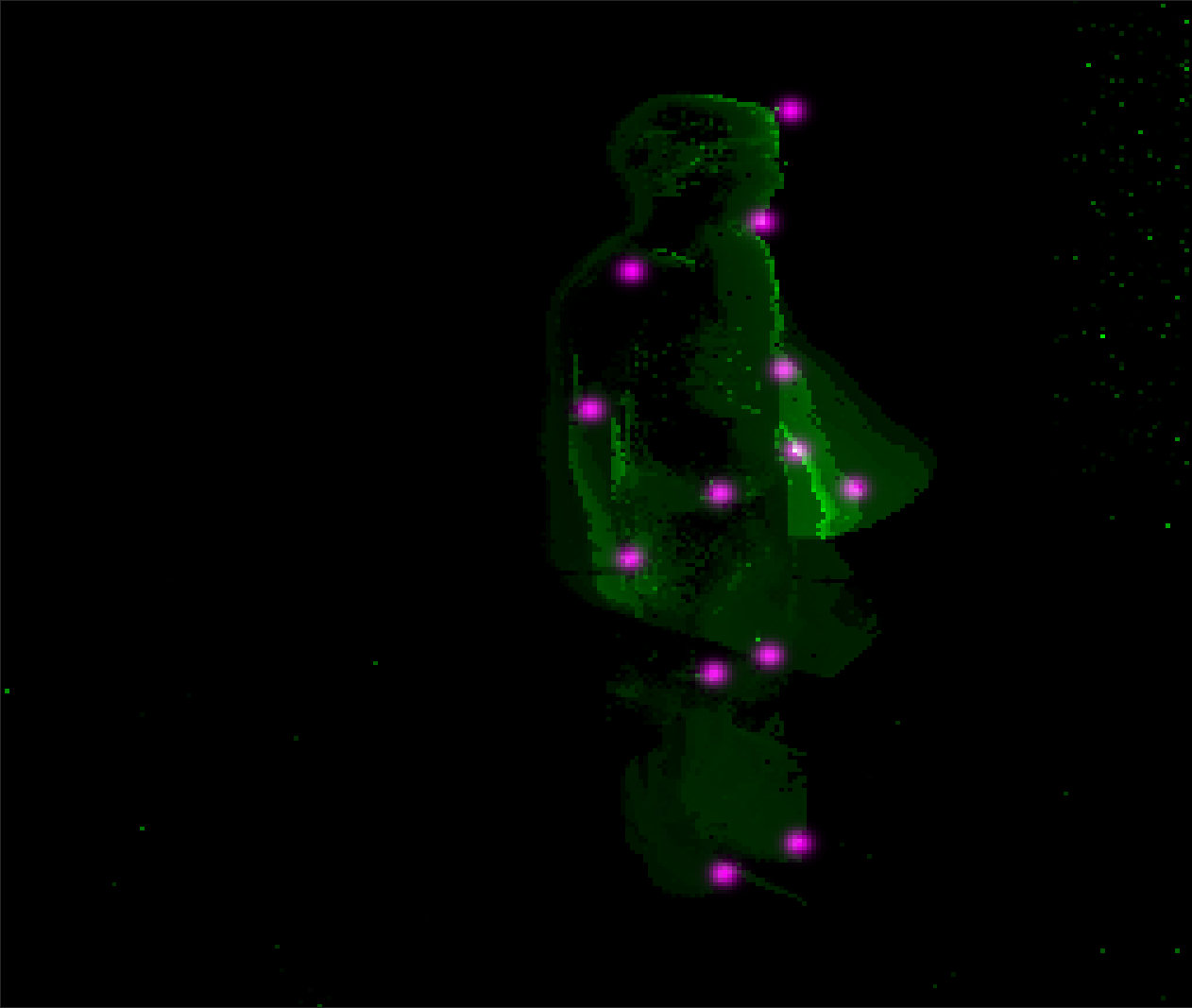} & \includegraphics[width=1.07\linewidth]{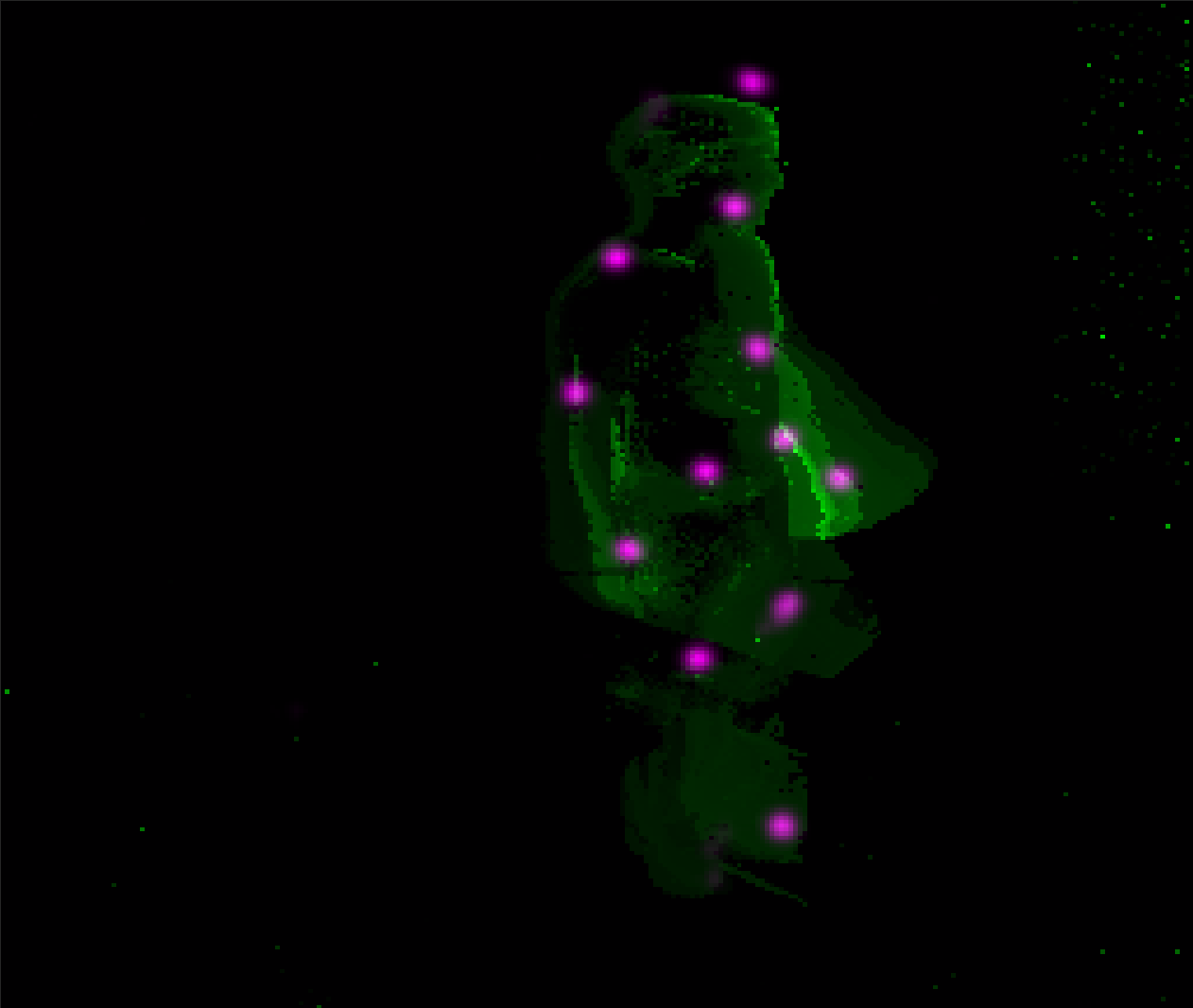}\\
\rotatebox{90}{Example 3} & \includegraphics[width=1.07\linewidth]{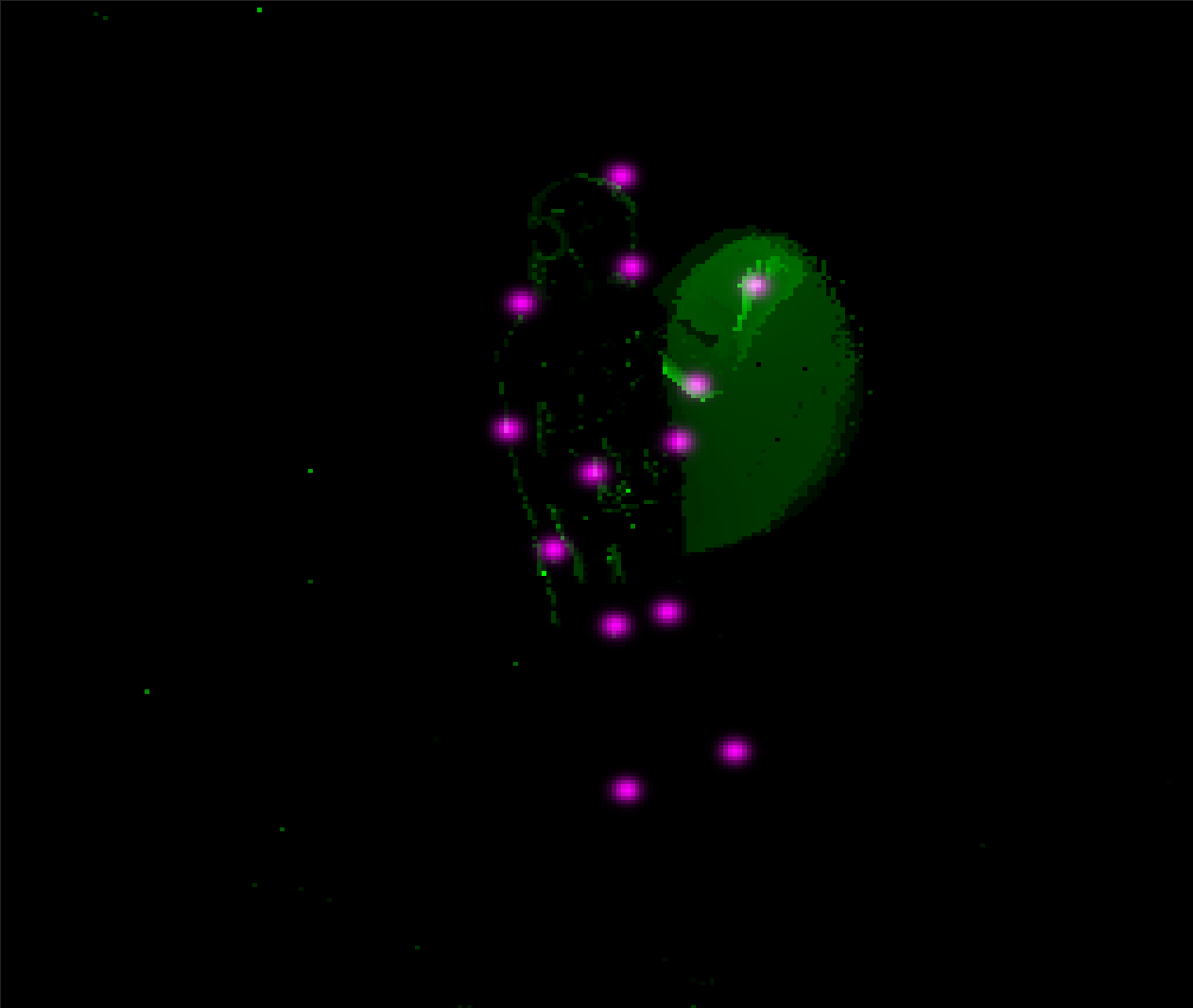} & \includegraphics[width=1.07\linewidth]{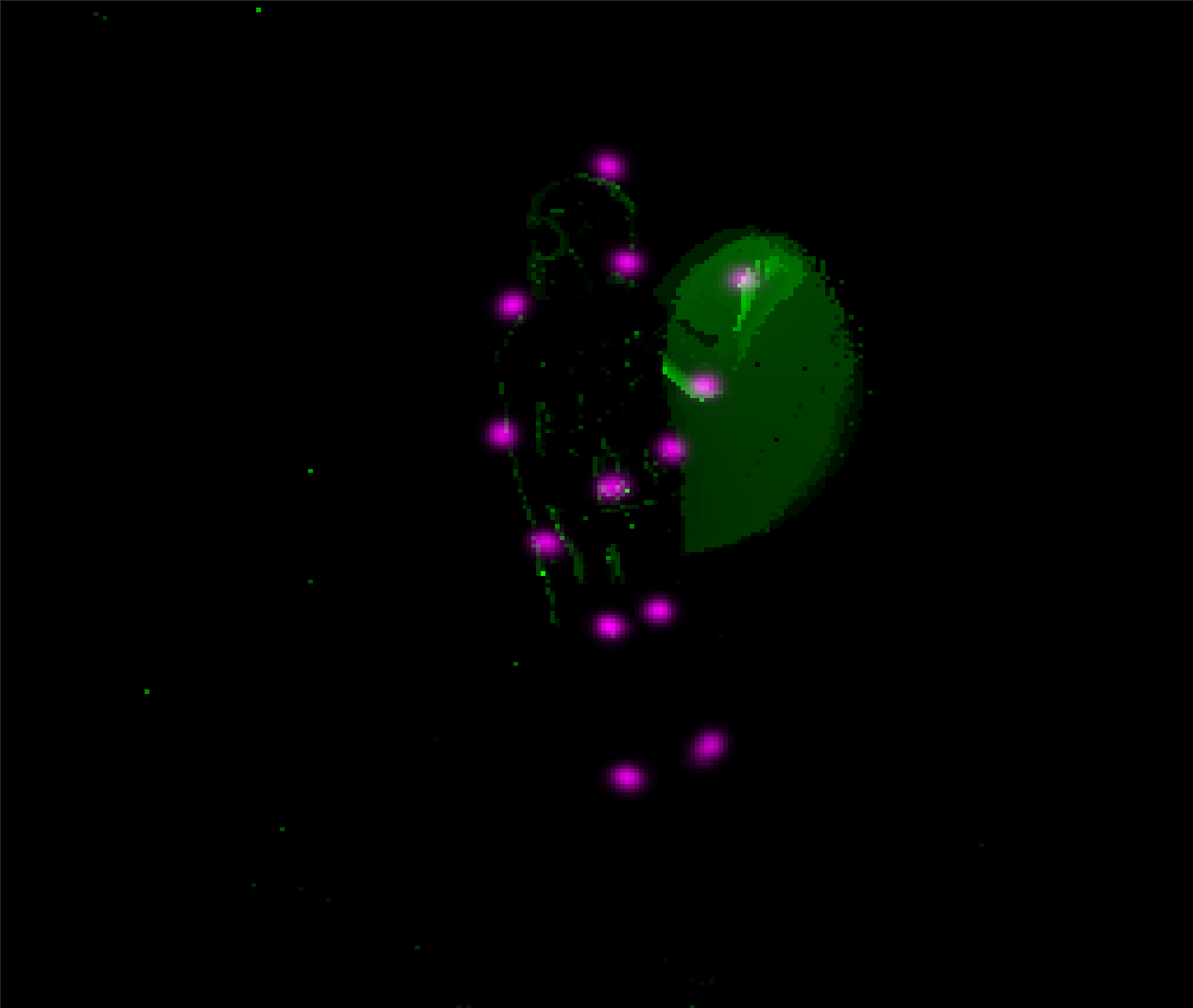}\\
\multicolumn{1}{r}{} & \multicolumn{1}{c}{Ground Truth} & \multicolumn{1}{c}{Proposed}
\end{tabular}
\caption{Qualitative 2D pose estimation results for different test subjects. Both images visualize the event representation in green and the 13 joint positions in pink. The left column shows the ground truth joint locations and the right column the network estimates from only event data.}
\label{fig:2dposesamples}
\end{figure}

We replicated all event camera data pre-processing steps for 2D pose estimation described in~\cite{calabrese2019dhp19}, including elimination of bad pixels, masking IR sources, and time-syncing 3D truth using special events within the data stream. The network architecture was also reused exactly. As such, the only difference between 2D pose estimation in~\cite{calabrese2019dhp19} and the proposed method is the event representation by TORE volumes in the latter. Figure~\ref{fig:leftHandTrack} shows the output results from the 2D pose estimation network. Even though no temporal smoothing has been applied, the TORE volume-based network output is smooth. This is a result of the temporal history embedded in the TORE event representation. Figure~\ref{fig:2dposesamples} overlays joint truth and predicted locations onto a single layer of the TORE volume.

Table~\ref{tab:2DMPJPE} shows the 2D results on the test set, expressed as mean per joint pixel error (MPJPE) in pixels for the two cameras used in prior work. The network was evaluated for prediction error both for instantaneous CNN prediction as well as for different values of confidence threshold, ranging from 0.1 to 0.5 as with prior work. Overall, the camera \#2 network using TORE volumes improved from an average per joint error of 7.18 to 5.07 and the camera \#3 network improved from 6.87 to 5.77.

\begin{table}
\centering
\caption{Test set 2D MPJPE (pixel) for the CNN, trained on the two frontal camera views (camera \#2 and \#3). In bold, the optimal threshold for each camera. Underlined, selected confidence threshold for 3D projection.}
\label{tab:2DMPJPE}
\scalebox{0.98}{
\begin{tabular}{cccccccc} 
\toprule
     &     & \multicolumn{6}{c}{Threshold}            \\ 
\cline{3-8}
Method & Cam & 0    & 0.1  & 0.2  & 0.3  & 0.4  & 0.5   \\ 
\hline\hline
CCF~\cite{calabrese2019dhp19}         & 2 & 7.72 & 7.45 & 7.25 & 7.18 & 7.27 & 7.47  \\
CCF~\cite{calabrese2019dhp19}         & 3 & 7.61 & 7.13 & 6.92 & 6.87 & 6.88 & 7.11  \\
\textbf{TORE (ours)} & 2 & 5.25 & \underline{\textbf{5.07}} & 5.08 & 5.24 & 5.62 & 9.08  \\
\textbf{TORE (ours)} & 3 & 5.98 & \underline{5.80} & \textbf{5.77} & 5.89 & 6.40 & 9.64  \\
\bottomrule
\end{tabular}}
\end{table}

\subsection{3D Pose Estimation}

The input to the 3D reconstruction network is a series of (estimated) 2D joint positions. In this work, 2D position was taken from the output of the TORE-based 2D joint position estimator described in Section \ref{sec:2dpose}, thresholded at value $0.1$ for cameras \#2 and \#3. As with prior work, 3D performance accuracy is measured using the standard evaluation metric MPJPE. MPJPE measures the 3D Euclidean distance between the network estimated joint positions and the ground truth positions in millimeters.

\begin{table}[ht]
    \begin{center}
    \scalebox{0.85}{
\begin{tabular}{cc|ccccc|c|cc}
\hline
\multicolumn{1}{c|}{}                                                                          &    & \multicolumn{5}{c|}{Testing Subject Number} & TORE & Prior         \\
\multicolumn{1}{c|}{Se}                                                                        & Mov  & Sub13   & Sub14  & Sub15  & Sub16  & Sub17  & Mean & SOTA            \\ \hline
\multicolumn{1}{c|}{\multirow{8}{*}{\begin{tabular}[c]{@{}c@{}}1\\ Mean\\ 58.4\end{tabular}}} & 1  & 27.34   & 38.64  & 39.39  & 62.17  & 130.56 & 59.6 & 112.8       \\
\multicolumn{1}{c|}{}                                                                          & 2  & 23.97   & 36.41  & 42.94  & 65.56  & 95.95  & 53.0 & 98.5        \\
\multicolumn{1}{c|}{}                                                                          & 3  & 30.71   & 133.21 & 41.42  & 61.13  & 61.12  & 65.5 & 84.6        \\
\multicolumn{1}{c|}{}                                                                          & 4  & 31.19   & 97.37  & 44.74  & 68.65  & 73.15  & 63.0 & 77.6        \\
\multicolumn{1}{c|}{}                                                                          & 5  & 30.84   & 43.29  & 43.66  & 86.71  & 113.50 & 63.6 & 103.2        \\
\multicolumn{1}{c|}{}                                                                          & 6  & 31.94   & 51.60  & 40.46  & 135.00 & 79.36  & 67.7 & 120.3        \\
\multicolumn{1}{c|}{}                                                                          & 7  & 23.50   & 42.91  & 39.94  & 63.16  & 61.78  & 46.3 & 74.5        \\
\multicolumn{1}{c|}{}                                                                          & 8  & 26.72   & 49.52  & 41.50  & 60.87  & 63.70  & 48.5 & 71.9        \\ \hline
\multicolumn{1}{c|}{\multirow{6}{*}{\begin{tabular}[c]{@{}c@{}}2\\ Mean\\ 47.9\end{tabular}}} & 9  & 22.84   & 47.91  & 32.61  & 61.33  & 50.89  & 43.1 & 53.6       \\
\multicolumn{1}{c|}{}                                                                          & 10 & 27.70   & 44.77  & 35.57  & 63.00  & 93.84  & 53.0 & 68.9        \\
\multicolumn{1}{c|}{}                                                                          & 11 & 38.29   & 44.74  & 38.10  & 67.21  & 75.67  & 52.8 & 74.2        \\
\multicolumn{1}{c|}{}                                                                          & 12 & 20.13   & 53.07  & 35.99  & 68.42  & 55.13  & 46.5 & 51.4        \\
\multicolumn{1}{c|}{}                                                                          & 13 & 34.69   & 45.42  & 40.72  & 62.13  & 53.23  & 47.2 & 54.8        \\
\multicolumn{1}{c|}{}                                                                          & 14 & 21.96   & 47.17  & 38.21  & 61.21  & 54.16  & 44.5 & 52.3        \\ \hline
\multicolumn{1}{c|}{\multirow{6}{*}{\begin{tabular}[c]{@{}c@{}}3\\ Mean\\ 70.5\end{tabular}}} & 15 & 28.08   & 151.43 & 53.28  & 85.53  & 62.63  & 76.2 & 144.8        \\
\multicolumn{1}{c|}{}                                                                          & 16 & 50.75   & 96.56  & 40.23  & 100.94 & 67.59  & 71.3 & 130.4        \\
\multicolumn{1}{c|}{}                                                                          & 17 & 37.03   & 137.83 & 42.84  & 77.92  & 60.79  & 71.3 & 107.1        \\
\multicolumn{1}{c|}{}                                                                          & 18 & 50.58   & 129.44 & 46.18  & 81.14  & 67.50  & 75.0 & 134.2        \\
\multicolumn{1}{c|}{}                                                                          & 19 & 27.44   & 116.23 & 42.38  & 77.93  & 72.79  & 67.4 & 100.6        \\
\multicolumn{1}{c|}{}                                                                          & 20 & 31.13   & 101.00 & 39.98  & 77.79  & 59.92  & 62.0 & 89.5        \\ \hline
\multicolumn{1}{c|}{\multirow{6}{*}{\begin{tabular}[c]{@{}c@{}}4\\ Mean\\ 62.8\end{tabular}}} & 21 & 25.23   & 49.41  & 46.23  & 67.13  & 55.26  & 48.7 & 54.8        \\
\multicolumn{1}{c|}{}                                                                          & 22 & 49.31   & 66.31  & 46.83  & 68.81  & 62.31  & 58.7 & 74.8        \\
\multicolumn{1}{c|}{}                                                                          & 23 & 35.49   & 80.42  & 64.53  & 63.47  & 66.42  & 62.1 & 109.4        \\
\multicolumn{1}{c|}{}                                                                          & 24 & 34.13   & 64.04  & 83.62  & 64.79  & 74.01  & 64.1 & 109.1        \\
\multicolumn{1}{c|}{}                                                                          & 25 & 45.76   & 75.26  & 96.84  & 62.56  & 87.55  & 73.6 & 115.4        \\
\multicolumn{1}{c|}{}                                                                          & 26 & 37.81   & 57.58  & 86.40  & 69.51  & 96.30  & 69.5 & 105.7       \\ \hline
\multicolumn{1}{c|}{\multirow{7}{*}{\begin{tabular}[c]{@{}c@{}}5\\ Mean\\ 53.0\end{tabular}}}  & 27 & 28.55   & 90.26  & 37.78  & 61.33  & 57.17  & 55.0 & 91.2        \\
\multicolumn{1}{c|}{}                                                                          & 28 & 26.71   & 47.12  & 38.26  & 69.09  & 62.93  & 48.8 & 94.0        \\
\multicolumn{1}{c|}{}                                                                          & 29 & 31.81   & *    & 38.14  & 68.18  & 71.51  & 52.4 & 107.8        \\
\multicolumn{1}{c|}{}                                                                          & 30 & 31.06   & 47.56  & 40.17  & 68.90  & 66.21  & 50.8 & 108.1        \\
\multicolumn{1}{c|}{}                                                                          & 31 & 36.48   & 42.45  & 38.13  & 67.15  & 64.47  & 49.7 & 111.0        \\
\multicolumn{1}{c|}{}                                                                          & 32 & 28.85   & 92.09  & 38.97  & 66.80  & 65.60  & 58.5 & 119.8        \\
\multicolumn{1}{c|}{}                                                                          & 33 & 27.55   & 50.57  & 57.18  & 78.34  & 63.68  & 55.5 & 124.5        \\ \hline
\multicolumn{2}{c}{Mean}                                                                       & 32.0    & 71.0   & 46.5   & 71.6   & 71.1   & \textbf{58.4} & 94.8 \\ \hline
\end{tabular}
}
    \end{center}
    \caption{3D pose reconstruction error for the DHP19 dataset. The table presents MPJPE (mm) for the five test subjects (Sub13 to Sub17) on all 33 movements (Mov). The final column is reported results from the prior SOTA on the DHP19~\cite{calabrese2019dhp19}.}
    \label{tb:MPJPE}
\end{table}

Prior work projected each 2D joint per camera into a ray in 3D space. The 3D estimated joint position was then calculated as the location that minimized distance to each ray. Our proposed method replaces this approach with a 2D to 3D pose estimation network that enhances temporal consistency and enables multi-camera feature fusion. The network employs an attention mechanism to provide a sequential prediction framework for learning spatial models. A large receptive field is crucial to learn long-range joint relationships across frames and achieve the highest accuracy. However, the number of neural layers and training parameters significantly increases with additional frames feeding the network. Therefore, a multi-scale dilation strategy of integrated dilated convolutions~\cite{liu2020attention} avoids large data redundancy.

\begin{figure}[htbp]
\centering
\begin{tabular}{m{0.01\linewidth}m{0.9\linewidth}}
\rotatebox{90}{R Knee Lift} & \includegraphics[width=1\linewidth]{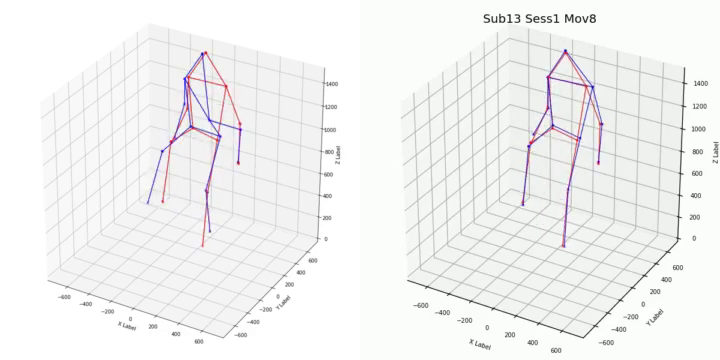}\\
\rotatebox{90}{R Leg Abduction} & \includegraphics[width=1\linewidth]{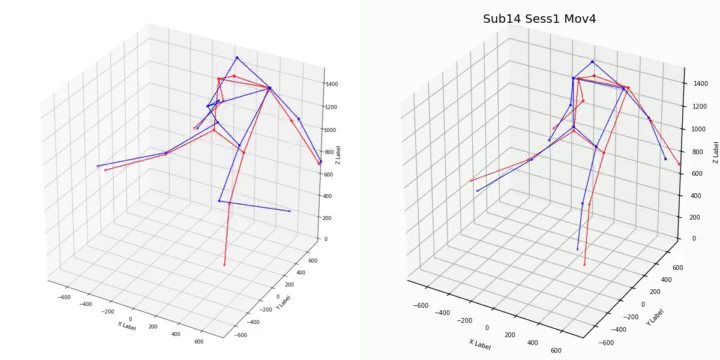}\\
\rotatebox{90}{Jump U/D} & \includegraphics[width=1\linewidth]{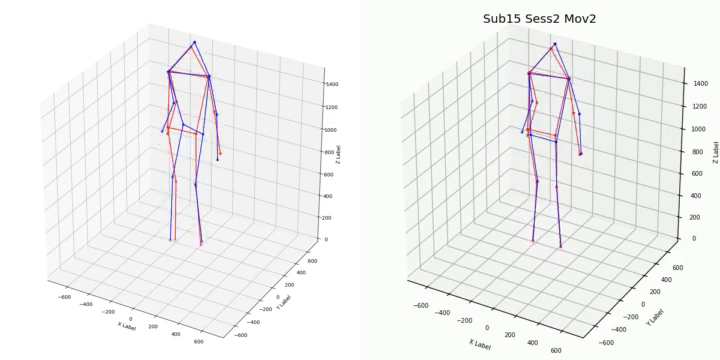}\\
\rotatebox{90}{Star Jump} & \includegraphics[width=1\linewidth]{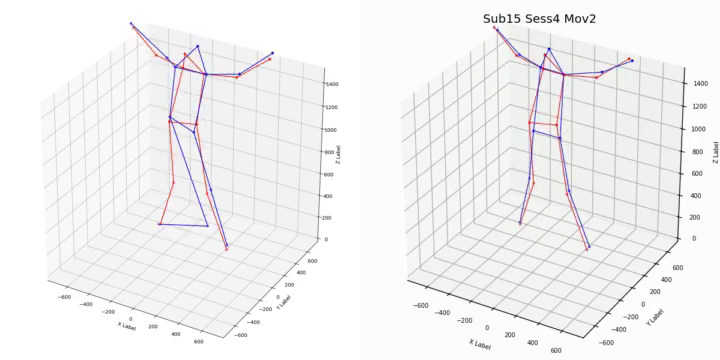}\\
\end{tabular}
\caption{(Left) Screenshots comparing ground truth 3D pose location (blue) to network estimated output (red) for four samples in the test dataset. Prior SOTA output (left) compared to proposed TORE-based network output (right). Animations are available via the project website.}
\label{fig:3dcompare}
\end{figure}

Table \ref{tb:MPJPE} reports the 3D MPJPE results for all subjects and movements together with averages across single subjects, movements, and sessions. Scores for nearly every session and movement were significantly better than prior state-of-the-art. The bold value highlights the overall mean error. Error was reduced by more than 36mm on average. As shown in Figure~\ref{fig:3dcompare}, the reconstruction quality improvement is easily visible when compared side-by-side to other reconstructions. 

\section{Conclusion}\label{sec:conclusion}

This paper presents Time-Ordered Recent Event (TORE) volumes, a novel representation of event camera data that arranges the past event time stamps in a FIFO format. It not only introduces zero latency, but is also easy to implement, adaptable for local or global tasks, and captures fine spatial and temporal details. We demonstrated that the proposed TORE volume representation helps achieve excellent performance in a wide range of challenging applications such as event denoising, image reconstruction, classification, and human pose estimation. TORE volume code and results can be found at the project GitHub site: \url{https://github.com/bald6354/tore_volumes}.

\ifCLASSOPTIONcompsoc
  \section*{Acknowledgments}
\else
  \section*{Acknowledgment}
\fi

This work was made possible in part by funding from Ford
University Research Program.

\ifCLASSOPTIONcaptionsoff
  \newpage
\fi



%
{\small
\bibliographystyle{IEEEtran}
\bibliography{egbib}
}
%
%

%





\begin{IEEEbiography}[{\includegraphics[width=1in,height=1.25in,clip,keepaspectratio]{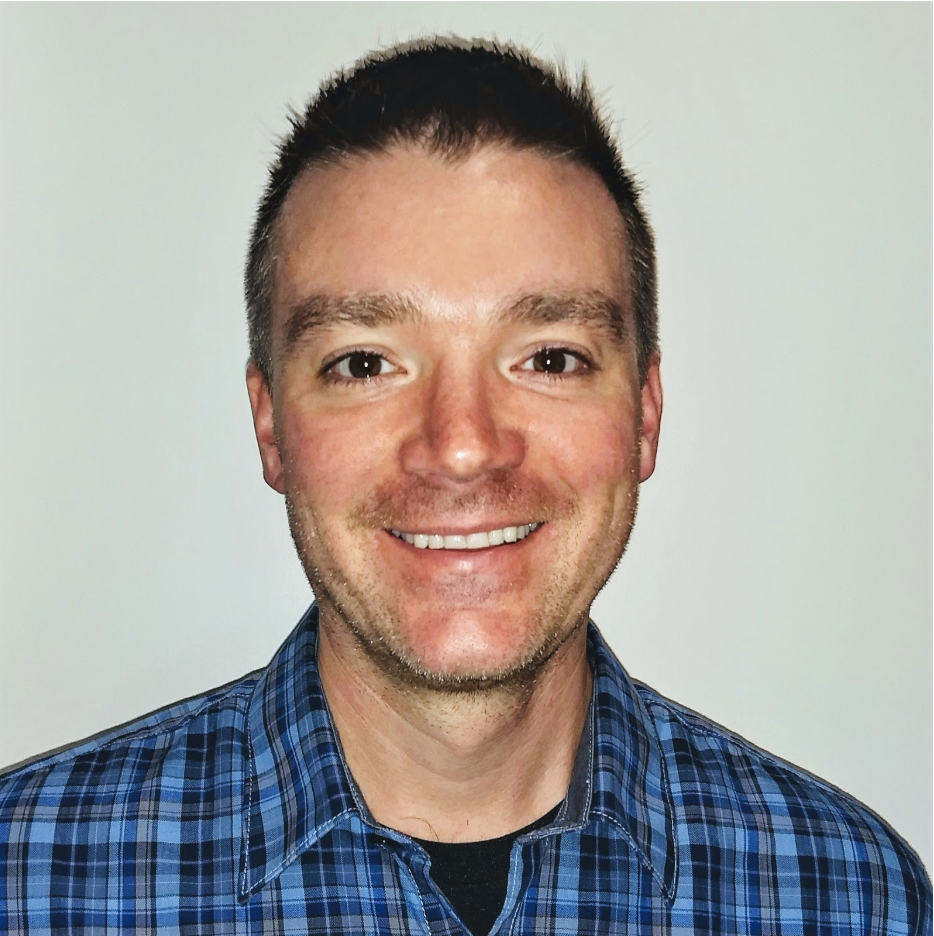}}]{R. Wes Baldwin}
received his B.S. in Computer Engineering in 2002 from Kettering University and his M.S. in Electrical and Computer Engineering in 2005 from the University of Illinois at Chicago. He is currently working towards a Ph.D. in Electrical and Computer Engineering at the University of Dayton. His research is focused on machine learning, image processing, and event cameras.
\end{IEEEbiography}

\begin{IEEEbiography}[{\includegraphics[width=1in,height=1.25in,clip,keepaspectratio]{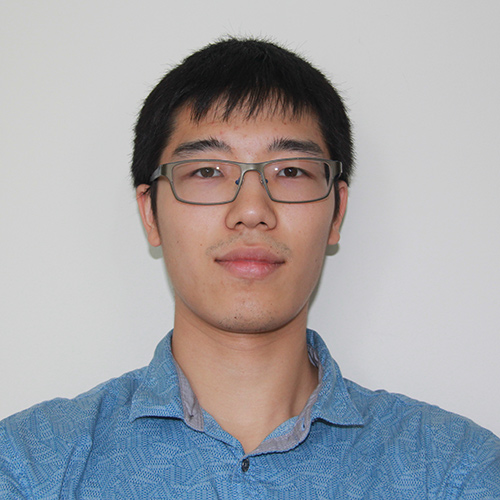}}]{Ruixu Liu}
received his Ph.D. and M.S. in Electrical and Computer Engineering from the University of Dayton, in 2019 and 2014 respectively. He received his B.S. degrees in Electrical Engineering from the Nanjing University of Science \& Technology in 2012. He is currently a research scientist at the University of Dayton, Ohio, USA. His research is focused on object detection, human pose estimation and 3D human pose reconstruction. 
\end{IEEEbiography}

\begin{IEEEbiography}[{\includegraphics[width=1in,height=1.25in,clip,keepaspectratio]{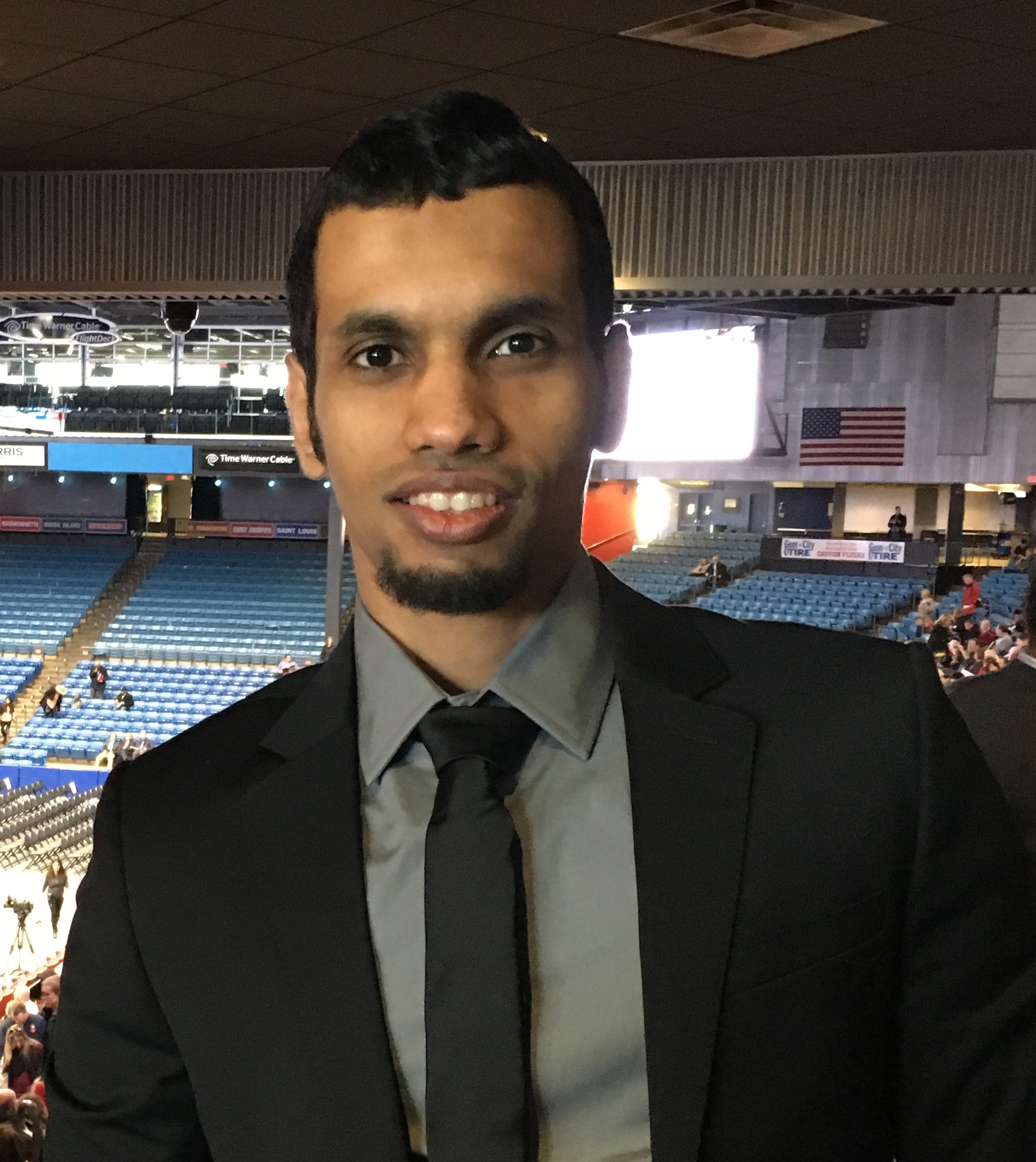}}]{Mohammed Almatrafi} 
(S’15-M’20) received his B.S. (Hons.) in Electrical Engineering from Umm Al-Qura University, Makkah, Saudi Arabia in 2011, the M.S. and Ph.D. degrees in Electrical and Computer Engineering from University of Dayton, Dayton, OH, USA, in 2015 and 2019, respectively. He is currently an Assistant Professor and the Vice Dean of the College of Engineering for Postgraduate Studies and Scientific Research at Umm Al-Qura University, Al-Lith, Saudi Arabia. His research is focused on  image processing, neuromorphic cameras and computer vision.   
\end{IEEEbiography}

\begin{IEEEbiography}[{\includegraphics[width=1in,height=1.25in,clip,keepaspectratio]{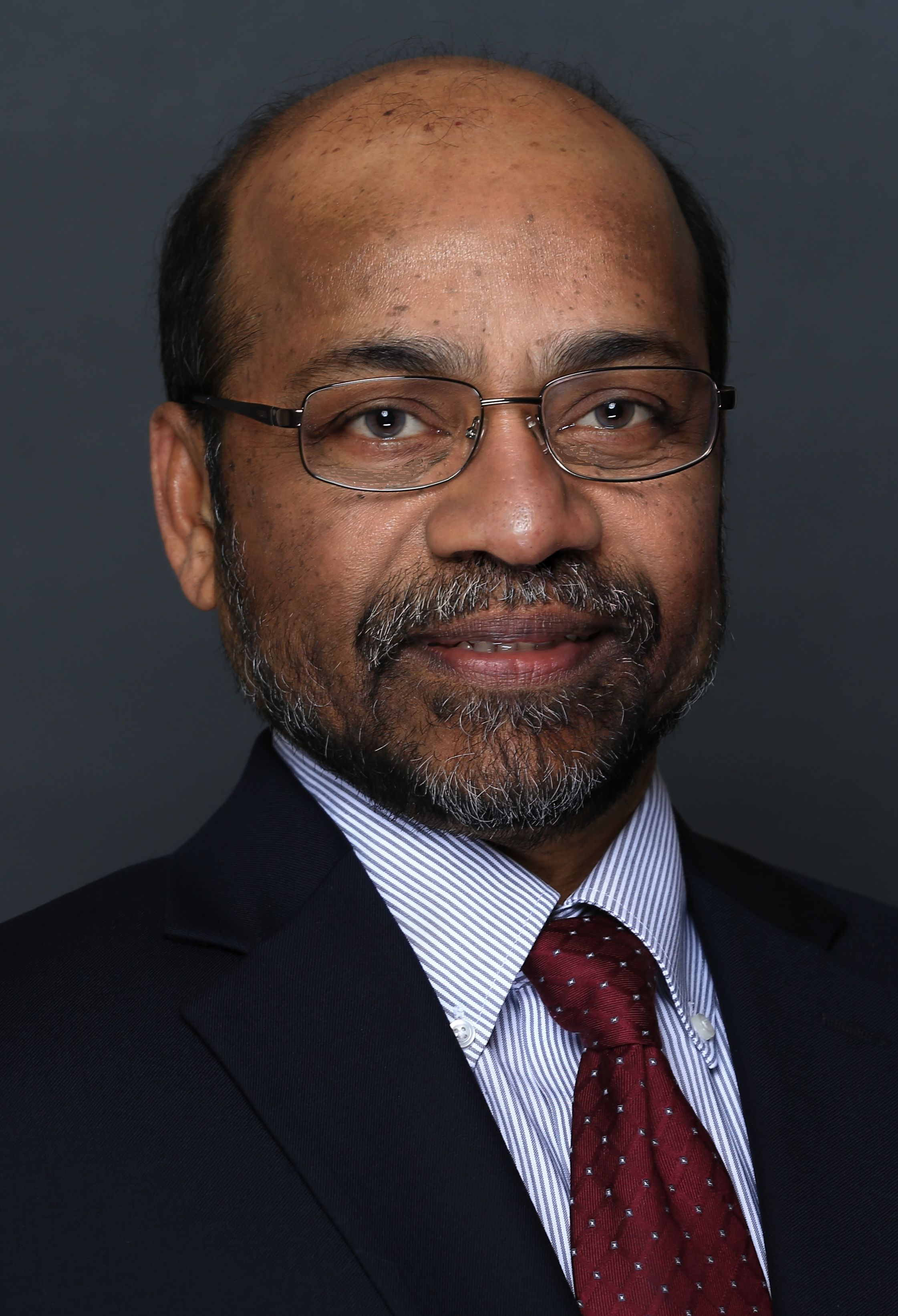}}]{Vijayan Asari}
Vijayan Asari received the Ph.D. degree in electrical engineering from the Indian Institute of Technology Madras in 1994. He is currently a Professor in electrical and computer engineering and Ohio Research Scholars Endowed Chair in wide area surveillance with the University of Dayton, Dayton, Ohio, USA. He is also the director of the Center of Excellence for Computational Intelligence and Machine Vision (Vision Lab) at UD. Dr. Asari holds four United States patents and has published more than 700 research articles including an edited book in wide area surveillance and 116 peer-reviewed journal papers in the areas of image processing, pattern recognition, machine learning, deep learning, and artificial neural networks. He is an elected Fellow of SPIE and a senior member of IEEE, and a co-organizer of several SPIE and IEEE conferences and workshops. Dr. Asari received several teaching, research, advising and technical leadership awards including the University of Dayton School of Engineering Vision Award for Excellence in August 2017, the Outstanding Engineers and Scientists Award for Technical Leadership from The Affiliate Societies Council of Dayton in April 2015, and the Sigma Xi George B. Noland Award for Outstanding Research in April 2016.
\end{IEEEbiography}

\begin{IEEEbiography}[{\includegraphics[width=1in,height=1.25in,clip,keepaspectratio]{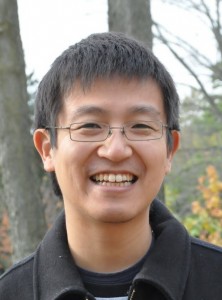}}]{Keigo Hirakawa}
(S’00–M’05–SM’11) received the
B.S. degree (Hons.) in electrical engineering from
Princeton University, Princeton, NJ, USA, in 2000,
the M.S. and Ph.D. degrees in electrical and computer engineering from Cornell University, Ithaca,
NY, USA, in 2003 and 2005, respectively, and the
M.M. degree (Hons.) in jazz performance studies
from the New England Conservatory of Music,
Boston, MA, USA, in 2006. He was a Research
Associate with Harvard University, Cambridge, MA,
USA from 2006 to 2009. He is currently an Associate Professor with the University of Dayton, Dayton, OH, USA. He is
currently the Head of the Intelligent Signal Systems Laboratory, University of
Dayton, where the group focuses on statistical signal processing, color image
processing, and computer vision.
\end{IEEEbiography}





\end{document}